\documentclass{article} 

\PassOptionsToPackage{table}{xcolor}

\usepackage{iclr2027_conference,times}

\usepackage[utf8]{inputenc}
\usepackage[T1]{fontenc}
\usepackage{hyperref}
\usepackage{url}
\usepackage{booktabs}
\usepackage{amsmath}
\usepackage{amssymb}
\usepackage{amsfonts}
\usepackage{nicefrac}
\usepackage{caption}
\usepackage{microtype}
\usepackage{xcolor}
\usepackage{graphicx}
\usepackage{xspace}
\usepackage{algorithm} 
\usepackage{algpseudocode} 
\usepackage{wrapfig} 
\usepackage{float} 
\usepackage{enumitem}

\newcommand{\ours}{\textsc{Staircase Policy}\xspace}

\newcommand{\ourmodel}{\textsc{S-WAM}\xspace}
\newcommand{\projecturl}{\url{https://s1ghhh.github.io/staircase-policy/}}

\newcommand{\fitwidth}[1]{%
 \resizebox{\ifdim\width>\linewidth\linewidth\else\width\fi}{!}{#1}}

\usepackage[normalem]{ulem}

\title{\ours: Streaming Inference for World-Action Models with Large Action Chunks}

\author{%
  \small
  \textbf{Guoheng Sun}$^{1}$
  \quad
  \textbf{Chen Chen}$^{2}$
  \quad
  \textbf{Jin Wang}$^{3}$
  \quad
  \textbf{Ang Li}$^{1}$
  \quad
  \textbf{Teresa Lv}$^{2}$\thanks{Corresponding author.}
  \\
  $^{1}$University of Maryland, College Park
  \quad
  $^{2}$Independent Researcher
  \\
  $^{3}$Oxford Robotics Institute, University of Oxford
}

\iclrfinalcopy 
\begin{document}

\maketitle
\lhead{}


\vspace{-0.2cm}

\begin{abstract}
World-Action Models (WAMs) improve robotic manipulation by conditioning action generation on predicted future observations, but future prediction adds further inference overhead to already expensive iterative action generation. Action chunking can amortize this cost over multiple actions, yet performance degrades over long execution horizons because later actions remain conditioned on stale observations. We introduce \ours{}, a streaming inference and training framework that turns a flow-matching VLA into a JEPA-style WAM and partitions a large action chunk into sub-chunks at staggered denoising stages. Near-term actions are executed as soon as they become available, while later actions continue to be refined. At each sub-chunk boundary, the future latent is re-predicted from the latest observation and used to update all unexecuted actions, enabling long-horizon execution without repeated full policy inference. The resulting future-prediction error can further serve as a signal for adaptive chunking.
\ourmodel{} achieves $97.7\%$ on LIBERO and $87.9\%$ on LIBERO-Plus, and improves performance across multiple policy backbones and real-robot tasks. It reaches $292.7$ executed actions per second, $3.62\times$ the throughput of conventional execution at comparable accuracy, while reducing time-to-first-action from $123.6$ to $73.3$\,ms. With additional inference optimizations, throughput further increases to $642.9$ actions per second.
The website is available at \projecturl{}.
\end{abstract}

\vspace{-0.2cm}

\section{Introduction}
\label{sec:intro}

\begin{wrapfigure}{r}{0.48\textwidth}
\vspace{-1cm}
\centering
\includegraphics[width=\linewidth]{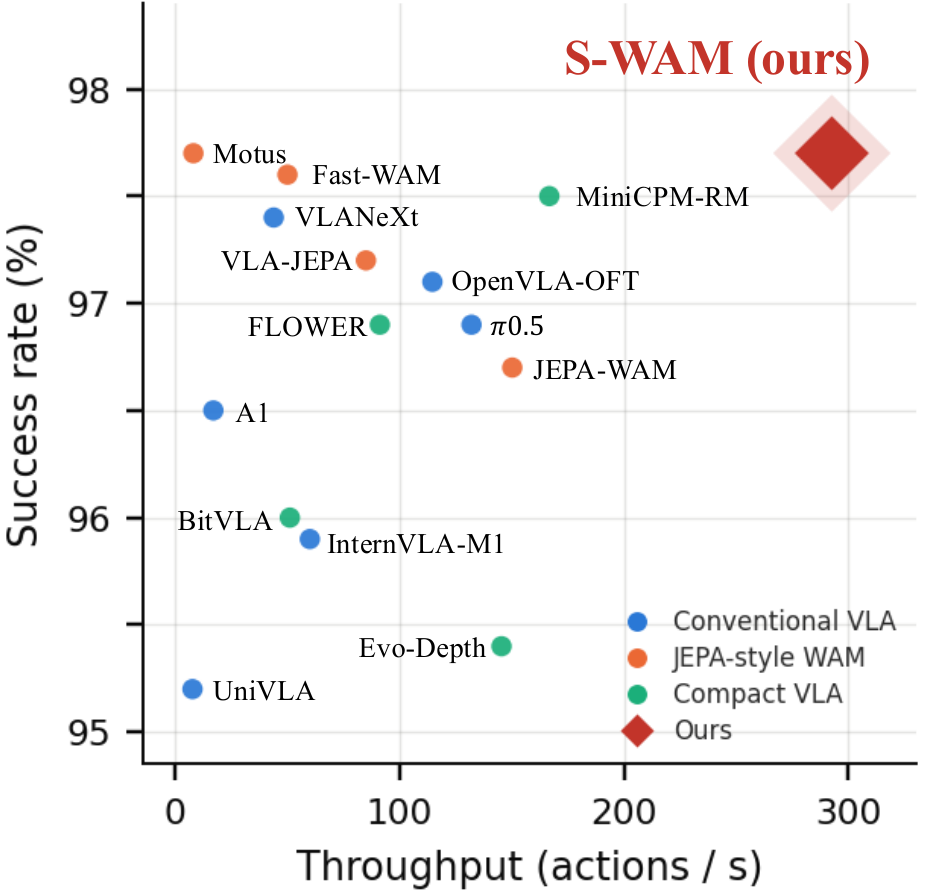}
\vspace{-0.4cm}
\caption{Success rate versus measured inference throughput on LIBERO. The measurement protocol, per-method results, and methods omitted from the figure are provided in Appendix~\ref{app:libero}.}
\label{fig:teaser}
\vspace{-0.6cm}
\end{wrapfigure}

Vision-language-action (VLA) models have achieved increasingly strong performance in robotic manipulation. These models typically combine a large pretrained vision-language backbone with an action decoder. Recent architectures often use a flow-matching-based Diffusion Transformer (DiT) as the action decoder to generate continuous actions directly~\citep{kim2024openvla,black2024pi0}. This design is becoming increasingly common in generalist manipulation policies and supports long action chunks in a single forward pass. 
A recent extension is the World-Action Model (WAM)~\citep{chen2026lawam,motubrain2026,streamvla2026}, which introduces an additional future-prediction module to predict a future observation or its latent representation.

Unlike WAMs that explicitly predict future images, JEPA-style variants predict a latent representation rather than pixels~\citep{jepavla2026,lin2026jepawam}. We study one instance of the latter family, where a lightweight predictor produces the future latent representation. Specifically, the current observation and the predicted future representation are then jointly used to condition action generation. Despite this more efficient design, future prediction still introduces additional inference overhead on top of an already expensive policy~\citep{chen2026lawam,yuan2026fastwam}. 
To sustain high-frequency control, the policy must generate new action chunks sufficiently frequently: a low-level controller may execute actions at several hundred hertz, while the policy inference frequency is typically below 10 Hz. 
A single WAM inference requires a forward pass through a multi-billion-parameter backbone and a future-prediction module, followed by iterative action denoising. 
This inference cost becomes particularly limiting for onboard deployment on edge devices~\citep{yang2026jetsonpi,tidal2026}. Improving the action throughput of WAMs without degrading task performance is therefore important for practical deployment.

Despite slow policy inference, action chunking provides a practical way to sustain high-frequency control~\citep{zhao2023act}. Given an observation $o_t$, the policy generates $H$ future actions in a single forward pass, allowing one expensive inference to be amortized over multiple control steps. Effective throughput, however, depends not on the generated horizon $H$, but on the executed horizon $H_{\mathrm{exec}} \leq H$: the number of generated actions actually executed before the next policy update. In practice, $H_{\mathrm{exec}}$ is often substantially smaller than $H$. A deployed system typically executes only the first few actions of a chunk, often one to twenty, acquires a new observation, queries the policy again, and discards the remaining actions~\citep{song2026faster}. 

This limited execution horizon is necessary because the reliability of later actions decreases as execution proceeds farther from the observation on which the chunk was conditioned. Tracking and contact errors accumulate while the robot acts without a new observation, uncertainty increases with the prediction horizon, and changes in the scene during execution are not reflected in the original conditioning observation. Increasing the generated chunk length alone therefore does not necessarily improve effective throughput. With the model weights and denoising budget fixed, increasing the executed horizon from $10$ to $50$ actions reduces success by $12.8$ points for LaWAM~\citep{chen2026lawam} and $26.8$ points for $\pi_{0.5}$~\citep{intelligence2025pi05} (Sec.~\ref{sec:disc:execlen}). The resulting challenge is to support a long execution horizon while allowing unexecuted actions to be continuously updated using the latest observation.

To address this challenge, we propose \ours, a streaming inference and training framework for World-Action Models with large action chunks: it adds a lightweight predictor of the future observation latent to a flow-matching VLA, turning it into the JEPA-style WAM we call an \ourmodel{}. \ours{} uses future prediction to maintain reliable actions over longer execution horizons. 
It maintains a buffer of sub-chunks at staggered denoising stages. 
First, it \textbf{progressively generates and immediately executes} near-term sub-chunks, allowing action execution to begin before the full chunk has been generated. Second, it \textbf{continuously refines unexecuted actions}: once a near-term sub-chunk has finished executing, a new observation is available, and \ours{} re-encodes that observation, re-predicts the future latent representation, and advances all remaining actions by one denoising step under the updated condition. 
This design supports long execution horizons while maintaining task performance: \ourmodel{} achieves $97.7\%$ on LIBERO and $87.9\%$ on LIBERO-Plus, with $292.7$ executed actions per second ($642.9$ with additional optimizations) and a $73.3$\,ms time-to-first-action (TTFA).
Our contributions are summarized as follows:
\begin{itemize}[leftmargin=1.2em, itemsep=2pt, topsep=2pt, parsep=0pt]
\item We propose \textbf{\ours, a streaming inference and training framework} for World-Action Models that pipelines the generation and execution of large action chunks. \ours{} maintains sub-chunks at staggered denoising stages, allowing near-term actions to be executed while later actions continue to be generated.

\item We introduce \textbf{observation-conditioned refinement} for long action chunks. As new observations arrive during execution, \ours{} updates the predicted future latent and continuously refines unexecuted actions, allowing the chunk to adapt without repeatedly invoking the full policy.

\item We show that \textbf{future-prediction error can serve as a practical signal for adaptive chunking}. By comparing the predicted future latent with the subsequently observed one, \ours{} can decide when to terminate the current chunk and replan.

\item We demonstrate \textbf{strong efficiency, robustness, and transferability} across multiple simulation benchmarks, policy backbones, and real-robot tasks. \ourmodel{} substantially improves action throughput and TTFA while maintaining strong task performance, including under perturbations and dynamic-scene evaluations.
\end{itemize}

\section{Related Work}
\label{sec:related}

\vspace{-0.05cm}

\textbf{World-action models.} Generalist manipulation policies combine a pretrained vision-language backbone with an iterative action decoder~\citep{chi2023diffusionpolicy,kim2024openvla,black2024pi0,intelligence2025pi05}. World-Action Models (WAMs) further condition actions on predicted future observations, either in pixel or video space~\citep{du2023unipi,streamvla2026,motubrain2026} or in representation space~\citep{jepavla2026,flare2025,ahead2026}, often using latent action modeling~\citep{bruce2024genie,chen2026lawam}. Future prediction adds inference overhead~\citep{chen2026lawam,yuan2026fastwam,yang2026jetsonpi}, and recent work questions whether it is needed at inference time~\citep{yuan2026fastwam,jepavla2026,wamrobust2026}. We instead study long execution horizons, where future representations are repeatedly updated to refine unexecuted actions.

\vspace{-0.05cm}

\textbf{Accelerating policy inference.} Prior work reduces iterative generation cost~\citep{luan2026snapflow,sants2026,kim2025openvlaoft}, trims redundant model capacity~\citep{sun2026dropthenrecovery}, overlaps inference with execution~\citep{black2025rtc,black2025ttrtc,asyncbench2026}, or streams generation with different noise levels across an action buffer~\citep{hoeg2024sdp,chen2024diffusionforcing,chen2025rnrdp,tidal2026,shi2026streamingvla}. Related methods asynchronously refresh semantic conditioning~\citep{park2026pir2}, train for intra-chunk inconsistency~\citep{remac2026}, stagger denoising over predicted video latents while sharing the action timestep~\citep{noisegate2026}, or separate slow semantic and fast action modules~\citep{chen2025fisvla}. FASTER gives each position of the chunk its own hit time so that the first action is ready after a single sampling step~\citep{song2026faster}. All of these condition a chunk on one observation and change only how it is generated. \ours{} keeps every position at a shared flow time, differing in readout depth alone, and re-conditions the unexecuted remainder on observations that arrive during execution.

\vspace{-0.05cm}

\textbf{Large action chunks.} Action chunking amortizes inference over multiple actions~\citep{zhao2023act}, yet published LIBERO policies typically execute only $1$--$20$ actions per query~\citep{kim2025openvlaoft,song2026faster,shi2026streamingvla,park2026pir2}, which caps the throughput benefit of chunking. Prior methods mitigate this by terminating execution early on uncertainty or prediction mismatch~\citep{feng2026dvac,wang2026ffdc,pan2026vlacorrector,ddp2026}, correcting chunks after generation~\citep{liu2024bid,sendai2025a2c2}, or conditioning generation on predicted future states~\citep{yang2026jetsonpi,ahead2026,pearlvla2026}. \ours{} instead continuously updates unexecuted actions as observations arrive, supporting a longer executed horizon without another full policy query.

\vspace{-0.2cm}

\section{Method}
\label{sec:method}

\ours{} is a streaming inference and training framework that turns a flow-matching VLA into a JEPA-style WAM. The design is motivated by two properties of long-horizon action generation.

\textbf{Denoising requirements vary across the action horizon.} Near-term actions are generally easier to predict, while actions farther into the future are less reliable~\citep{song2026faster}. This suggests allocating denoising computation non-uniformly across the chunk. Because flow-matching trajectories are approximately straight~\citep{liu2023rectifiedflow,lipman2023flowmatching}, a single velocity evaluation can already provide a useful estimate toward the data endpoint. \ours{} therefore emits the first sub-chunk after one denoising step and gives each subsequent sub-chunk one additional step (Sec.~\ref{sec:method-staircase}).

\textbf{Visual conditioning should be refreshed during execution.} Object poses, contacts, and robot configuration change as actions are executed, while the task instruction and high-level semantic context remain largely unchanged within a chunk. \ours{} therefore refreshes only the lightweight vision-side future predictor at each sub-chunk boundary, while reusing the cached vision-language context. This provides updated visual conditioning at millisecond-scale cost without repeatedly invoking the substantially more expensive backbone (Sec.~\ref{sec:method-refresh}).

\begin{figure}[t]
\centering
\includegraphics[width=\linewidth]{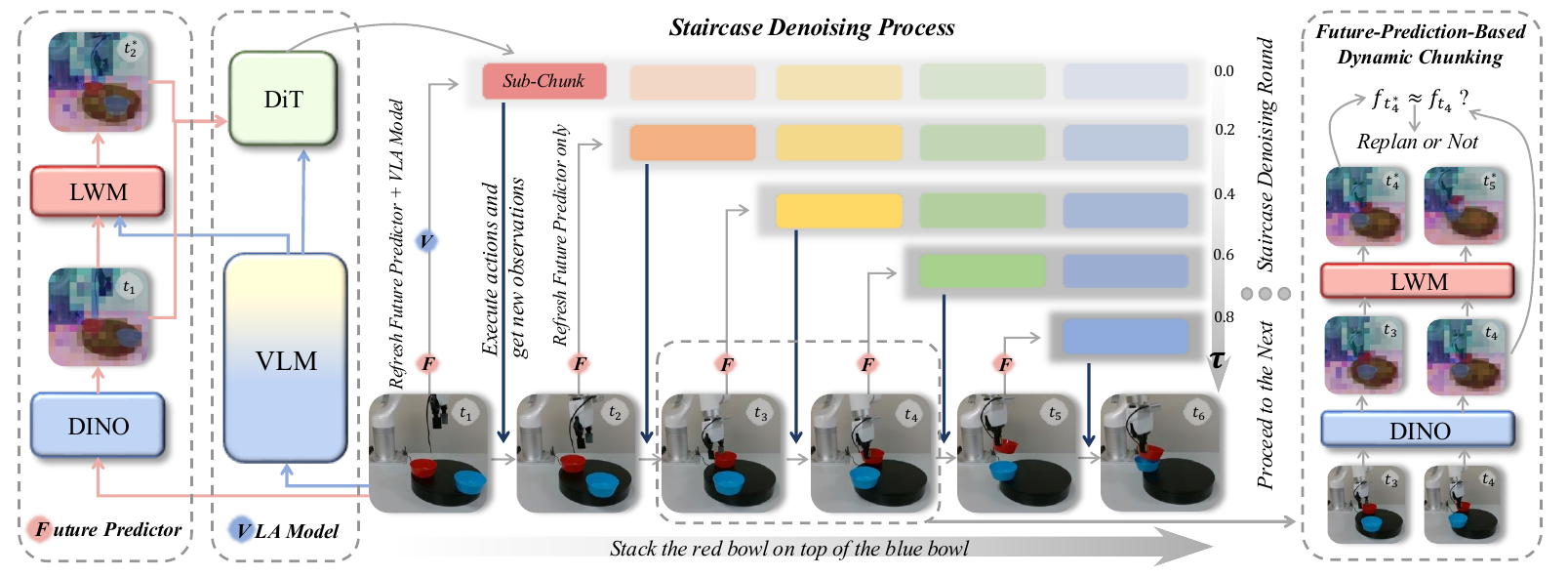}
\caption{Overview of \ours{}. Here $t_i$ denotes a time step and $t_i^{*}$ the predicted future.}
\label{fig:method}

\end{figure}

\subsection{Setup}
\label{sec:method-prelim}

The host is a flow-matching VLA mapping an observation $o$ and instruction $\ell$ to a chunk of $H$ actions~\citep{zhao2023act}: a \textbf{backbone} of billions of parameters emits a semantic context $e$ and a compact latent plan $z$~\citep{bruce2024genie}, and a flow-matching \textbf{expert} $v_\theta$ denoises a buffer $x \in \mathbb{R}^{H \times d_a}$ toward the data along the linear path from noise~\citep{lipman2023flowmatching}. We require only that the expert be callable as a \emph{single} Euler step against a cached backbone context.

\ours{} adds a lightweight \textbf{future predictor} consisting of a frozen vision encoder $h = E(o)$ and a decoder $g$ that predicts the feature map of a near-future observation, the \emph{future latent} $\hat{f} = g(h, z)$. The expert is conditioned on $c = (h, \hat{f}, e)$, and we refer to the resulting streaming World-Action Model as an \ourmodel{}. The future predictor we introduce follows the latent world model of LaWAM~\citep{chen2026lawam}, retaining its predictor architecture while training it under the streaming schedule described below.

Under conventional inference, the entire action buffer is denoised for $n$ steps, after which the policy executes a prefix of $H_{\mathrm{exec}} \le H$ actions without taking a new observation and discards the remainder. 
Increasing $H_{\mathrm{exec}}$ improves executed-action throughput, but also requires executing later actions, which are less reliable as execution moves farther from the conditioning observation. 
\ours changes how the action chunk is denoised and executed: instead of fully denoising the entire chunk before execution, it progressively generates near-term actions while continuously updating the remaining actions with new observations, without modifying the host backbone or flow expert.

\begin{wrapfigure}{r}{0.52\textwidth}
\vspace{-0.9cm}
\begin{minipage}{0.52\textwidth}
\begin{algorithm}[H]
\caption{\ours{} inference for one chunk of $H = KG$ actions}
\label{alg:streamwam}
\begin{algorithmic}[1]
\State \textbf{once per chunk:} $(e, z) \gets \mathrm{backbone}(o_0, \ell)$
\State $x \sim \mathcal{N}(0, I)$;\quad $\tau \gets 0$
\For{$j = 0, \dots, K-1$}
 \State $h_j \gets E(o_j)$;\quad $\hat{f}_j \gets g(h_j, z)$
 \Comment{refresh, Eq.~\eqref{eq:refresh}}
 \State $v \gets v_\theta(x, \tau \mid h_j, \hat{f}_j, e)$
 \Comment{one expert pass}
 \State emit $\hat{A}_j = [\,x + (1-\tau)\,v\,]_{jG:(j+1)G}$
 \Comment{readout, Eq.~\eqref{eq:readout}}
 \State $x \gets x + \frac{1}{K}v$;\quad $\tau \gets \tau + \frac{1}{K}$
 \State execute $\hat{A}_j$ ($G$ steps); observe $o_{j+1}$
\EndFor
\end{algorithmic}
\end{algorithm}
\end{minipage}
\vspace{-0.0cm}
\end{wrapfigure}

\subsection{The Denoising Staircase}
\label{sec:method-staircase}

We partition the chunk into $K$ contiguous sub-chunks of size $G$, so that the chunk length is $H = KG$. These two hyper-parameters are the only scheduling parameters varied in our study (Sec.~\ref{sec:disc:kgran}); unless otherwise specified, we use $H{=}50$, $K{=}5$, $G{=}10$. Algorithm~\ref{alg:streamwam} states the resulting inference loop.

The buffer starts as pure noise at a single shared flow time $\tau{=}0$, and each query advances \emph{all} positions by one Euler step of size $1/K$, $x \leftarrow x + \frac{1}{K} v_\theta(x, \tau \mid c_j)$ with $\tau \leftarrow \tau + \frac{1}{K}$, giving the grid on the right of Fig.~\ref{fig:method}. Immediately after the $j$-th update ($j = 0,\dots,K{-}1$) the $j$-th sub-chunk is \emph{read out} by extrapolating the same velocity estimate to $\tau{=}1$,
\begin{equation}
 \hat{A}_j = \big[\, x + (1-\tau)\, v_\theta(x, \tau \mid c_j) \big]_{\,jG:(j+1)G},
 \label{eq:readout}
\end{equation}
and handed to the controller while the rest of the buffer keeps denoising. We denote the full extrapolated buffer before slicing as $\hat{A}_j^{\,\mathrm{full}}$.

Sub-chunk $j$ is therefore emitted after $j{+}1$ forward passes of the DiT action expert. Although different sub-chunks have different readout depths, all buffer positions remain at the same flow time $\tau$, so the expert always receives inputs from the shared-$\tau$ distribution used during training. Per $H$ executed actions, the schedule requires one backbone pass and $K$ denoising steps.

\subsection{Refreshing the Future Predictor}
\label{sec:method-refresh}

The denoising staircase creates $K$ sub-chunk boundaries within each chunk. At every boundary, a new observation is available when the expert performs its next denoising update. After completing sub-chunk $j{-}1$, the robot observes $o_j$, and \ours{} recomputes the future prediction as
\begin{equation}
 h_j = E(o_j), \quad \hat{f}_j = g(h_j, z), \quad c_j = (h_j,\, \hat{f}_j,\, e),
 \label{eq:refresh}
\end{equation}
while keeping $z$ and $e$ fixed at their chunk-start values. Thus, the expensive VLA backbone runs one forward pass per full chunk, whereas the lightweight future predictor runs once per sub-chunk boundary, as illustrated in Fig.~\ref{fig:method}. 
Each refreshed condition is applied to all unexecuted positions in the buffer. Consequently, changes observed at boundary $j$ can influence every subsequent action in the current chunk without requiring a full policy replan. This observation-conditioned refresh is therefore what allows the staircase schedule to maintain action quality over a long execution horizon.

Because the refresh runs as a chain, the predicted future is itself a usable signal: $\hat{f}_{j-1}$ targets $h_j$, which is encoded one sub-chunk later, so every boundary yields a prediction--realization pair that a policy denoising the chunk only once never has. Their discrepancy $\delta_j = \operatorname{MSE}(\hat{f}_{j-1},\, h_j)$ grows when the scene departs from what the model anticipated at planning time, for instance because the target was displaced. If $\delta_j$ exceeds a threshold, the chunk ends after sub-chunk $j$ and the backbone plans a new one (right of Fig.~\ref{fig:method}), making the executed horizon data-dependent rather than fixed at $K$ sub-chunks. 
We explore such signal in detail in Sec.~\ref{sec:disc:signal}.

\subsection{Training Under the Deployment Schedule}
\label{sec:method-training}

 Standard training does not expose the policy to several states encountered by Algorithm~\ref{alg:streamwam}, including partially denoised buffers, one-step readouts, changes in conditioning within a chunk, and self-predicted future latents. We therefore train the model using the same streaming schedule used at inference time. We further find that training this way converges considerably faster than conventional chunk training (Sec.~\ref{sec:exp:budget}). Each demonstration window provides the $K{+}1$ observations at the sub-chunk boundaries. At every boundary, the condition uses the \emph{self-predicted} $g(h_j, z)$ rather than the ground-truth $h_{j+1}$, matching the information available during deployment.

The forward rollout remains sequential, while the action buffer is detached between boundaries to avoid backpropagation through the full $K$-step chain. The $K$ expert graphs are optimized in a shared backward pass.

Supervision targets the quantity the controller consumes, the readout~\eqref{eq:readout} rather than the velocity. At boundary $j$ each buffer position carries weight $1$ if it lies in the emitted sub-chunk and $\lambda_{\mathrm{ne}}{=}0.25$ otherwise, and the losses pool boundaries:
\begin{equation}
 \mathcal{L}_{\mathrm{act}}
 = \frac{\sum_{j,p} w_j(p)\, \| \hat{A}_j^{\,\mathrm{full}}(p) - A(p) \|_2^2}
 {d_a \sum_{j,p} w_j(p)},
 \qquad
 \mathcal{L}_{\mathrm{f}}
 = \frac{1}{K} \sum_{j=0}^{K-1} \operatorname{MSE}\!\left(g(h_j, z),\, h_{j+1}\right) .
 \label{eq:losses}
\end{equation}

Executed positions receive full weight, while the lower-weight dense term supervises all other valid positions in the buffer. The future latent remains attached to the computation graph: gradients from $\mathcal{L}_{\mathrm{act}}$ propagate through $\hat{f}_j$ into $g$ and $z$. Thus, the future predictor is optimized both for latent prediction and for its utility in action generation. The total objective is $\mathcal{L} = \mathcal{L}_{\mathrm{act}} + \beta(\mathcal{L}_{\mathrm{f}} + \mathcal{L}_{z})$ with $\beta = 0.1$, where $\mathcal{L}_{z}$ distills $z$ toward the frozen latent-action encoder of the pretraining stage.

\section{Experiments}
\label{sec:experiments}

\subsection{Experimental Setup}
\label{sec:exp:setup}

\paragraph{Simulation Benchmarks.}
We evaluate on three simulation benchmarks: \textsc{LIBERO}~\citep{liu2023libero}, with four suites of ten tasks each; \textsc{LIBERO-Plus}~\citep{liberoplus2025}, with seven perturbation categories over the same tasks; and \textsc{DOMINO}~\citep{domino2026}, a dynamic-manipulation benchmark with scene changes during execution. For all benchmarks, we jointly train a single model over the full set of constituent tasks. For example, on \textsc{LIBERO}, all four suites are mixed during training.

\paragraph{Model and Training Details.}
Unless otherwise specified, we use a $2.3$B latent World-Action Model as the base VLA. Within each backbone, \ourmodel{} and the vanilla baseline use the same data, initialization, and training configuration; full training recipes are provided in Appendix~\ref{app:setup-train}. The vanilla policy executes a fixed prefix of $H_{\mathrm{exec}}$ actions before replanning. For every $50$ executed actions, \ourmodel{} with $K=5$ and vanilla execution with $H_{\mathrm{exec}}=50$ both require five denoiser passes and one backbone pass, while $H_{\mathrm{exec}}=10$ requires $25$ and five. We therefore use $H_{\mathrm{exec}}=50$ as the compute-matched baseline and $H_{\mathrm{exec}}=10$ as the conventional short-horizon baseline. On \textsc{LIBERO}, each task is evaluated over $50$ trials; \textsc{LIBERO-Plus} uses its full test set.

\subsection{Real-Robot Experiments}
\label{sec:exp:real}

\begin{figure}[t]
\centering
\includegraphics[width=\linewidth]{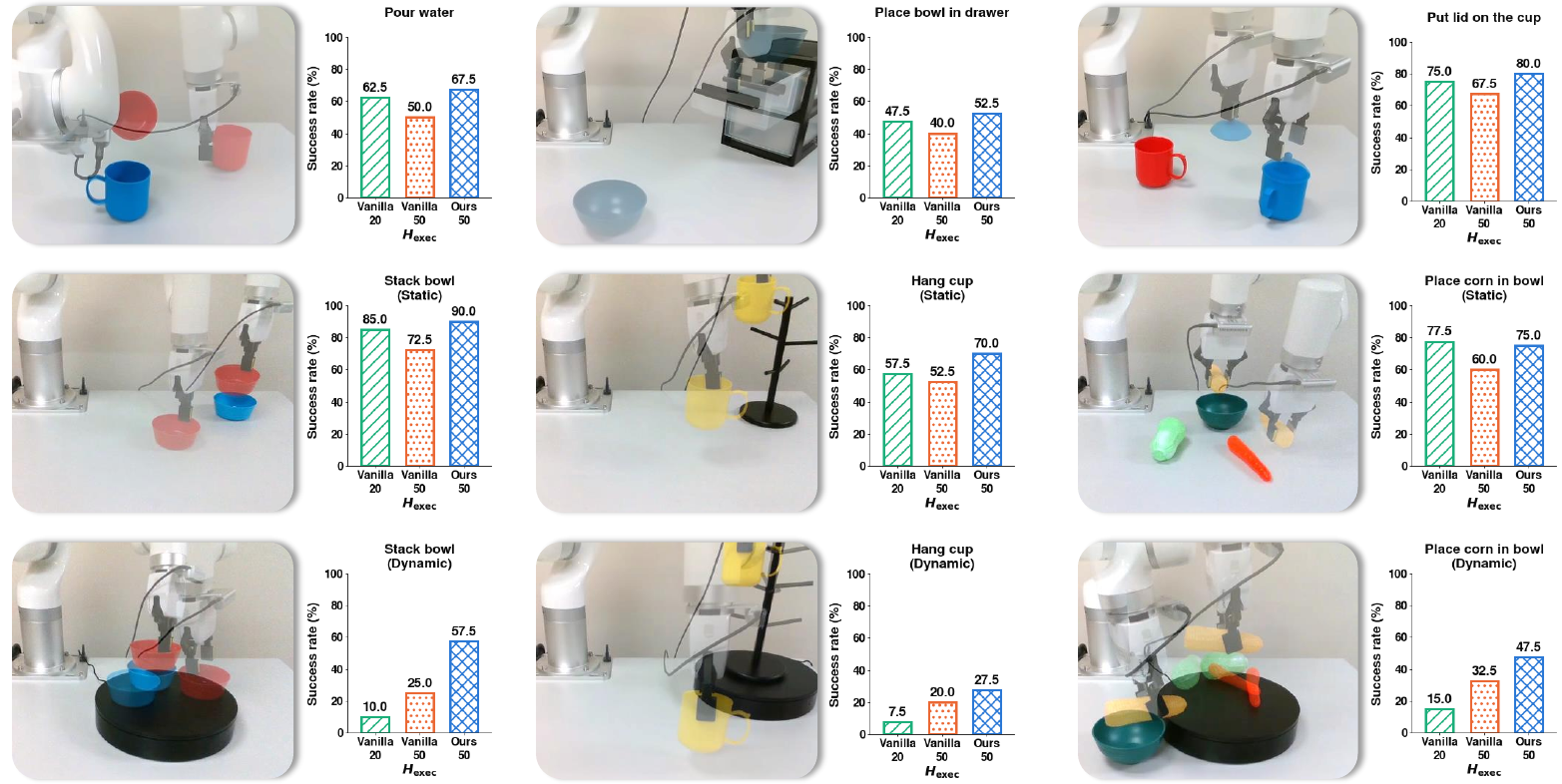}
\caption{Real-robot results on nine tasks. Task settings and details are provided in Appendix~\ref{app:real-datasets}.}
\label{fig:real}
\vspace{-0.5cm}
\end{figure}

\paragraph{Setup.}
We evaluate nine real-robot tasks on a UFACTORY xArm~850, with the policy running onboard an NVIDIA Jetson Thor. For each task we collect $30$ teleoperated demonstrations using a Meta Quest~3. Six of the nine are static and dynamic versions of three manipulation skills, counted separately; in the dynamic version the target objects move continuously on a rotating turntable. The remaining three cover sustained, precise, and multi-stage manipulation. Each task is evaluated over $40$ trials. Dataset statistics, observation and action spaces, scene randomization, success criteria, and training details are provided in Appendix~\ref{app:real-datasets} and Appendix~\ref{app:setup-train}.

\paragraph{Results.}
As shown in Fig.~\ref{fig:real}, on the six static tasks \ourmodel{} reaches $72.5$ success, compared with $67.5$ for vanilla at $H_{\mathrm{exec}}{=}20$ and $57.1$ at $H_{\mathrm{exec}}{=}50$. The drop at $H_{\mathrm{exec}}{=}50$ reflects the limitation of vanilla large-chunk execution over long horizons. The dynamic tasks expose the limitations of both vanilla execution regimes. Frequent replanning is affected by inference latency while the scene continues to move, yielding only $10.8$ at $H_{\mathrm{exec}}{=}20$. Executing $50$ actions per query improves this to $25.8$, but the later actions in the large chunk accumulate prediction drift over the long execution horizon. \ourmodel{} instead updates the unexecuted actions from new observations and reaches $44.2$. Detailed per-task results are provided in Table~\ref{tab:app-real}.

\subsection{Performance of \ourmodel{} Across Simulation Benchmarks}
\label{sec:exp:main}

\paragraph{LIBERO.} Figure~\ref{fig:teaser} compares \ourmodel{} with $14$ released policies in terms of Success Rate (SR) and measured throughput. Existing methods exhibit a clear speed--accuracy trade-off: policies with higher SR often rely on shorter replanning intervals and larger backbones, resulting in lower throughput, while faster policies typically achieve lower SR. 
\ourmodel{} reaches $97.7\%$ SR at $292.7$ executed actions per second, matching the highest SR while achieving $1.8\times$ the throughput of the next-fastest method shown. SRs are taken from the original papers, while throughput numbers are measured from released checkpoints under a unified protocol. 
Full results and details are provided in Appendix~\ref{app:libero}.

\paragraph{LIBERO-Plus.} Table~\ref{tab:plus} evaluates robustness across seven perturbation dimensions. \ourmodel{} achieves the best average SR, $87.9\%$, outperforming the vanilla policy at $H_{\mathrm{exec}}{=}50$ by $17.3$. Robot initial-state perturbation is where that baseline is weakest ($49.7$); \ourmodel{} reaches $76.0$ there, the best score in the table, consistent with the benefit coming from re-observing scene geometry during execution. Compared with vanilla execution at $H_{\mathrm{exec}}{=}10$, \ourmodel{} achieves comparable overall SR. The per-suite breakdown is given in Appendix~\ref{app:libero-plus-detailed}.


\begin{table}[t]
\centering
\caption{Robustness on LIBERO-Plus across seven perturbation dimensions.}
\label{tab:plus}
\setlength{\tabcolsep}{4pt}
\fitwidth{%
\begin{tabular}{lrrrrrrrr}
\toprule
Method & Background & Robot & Camera & Language & Noise & Layout & Light & Avg. \\
\midrule
ST-WAM~\citep{wang2026stwam} & 74.2 & 60.1 & 55.4 & 79.3 & 79.5 & 74.3 & 93.0 & 73.7 \\
DreamWAM~\citep{yuan2026dreamwam} & 71.5 & 63.6 & 53.7 & \textbf{94.8} & 67.1 & 80.7 & 96.6 & 75.4 \\
VLA-JEPA~\citep{sun2026vlajepa} & 93.6 & 67.1 & 63.3 & 85.4 & 66.3 & 85.1 & 95.6 & 79.5 \\
ROCKET-VLA~\citep{sun2026rocket} & 91.8 & 41.8 & \textbf{91.8} & 78.0 & 92.5 & 81.2 & 94.7 & 81.7 \\
VLANeXt~\citep{wu2026vlanext} & 82.5 & 65.7 & 90.4 & 81.8 & 94.1 & 80.8 & 95.9 & 84.5 \\
WorldPilot~\citep{lin2026worldpilot} & 96.4 & 60.6 & 82.8 & 87.2 & 93.6 & 80.5 & \textbf{98.6} & 85.7 \\
InternVLA-A1.5~\citep{ma2026internvlaa15} & \textbf{98.2} & 55.1 & 83.1 & 86.9 & \textbf{95.6} & \textbf{85.2} & 96.4 & 85.8 \\
\midrule
Vanilla, $H_{\mathrm{exec}}{=}50$ & 83.0 & 49.7 & 77.1 & 53.9 & 80.0 & 64.8 & 85.7 & 70.6 \\
Vanilla, $H_{\mathrm{exec}}{=}10$ & 95.2 & 73.4 & 91.5 & 77.7 & 94.3 & 82.1 & 96.0 & 87.2 \\
\ourmodel{} (ours), $H_{\mathrm{exec}}{=}50$ & 96.6 & \textbf{76.0} & 89.5 & 85.2 & 90.4 & 79.3 & \textbf{98.6} & \textbf{87.9} \\
\bottomrule
\end{tabular}
}

\end{table}


\begin{figure}[htbp]
\centering
\includegraphics[width=\linewidth]{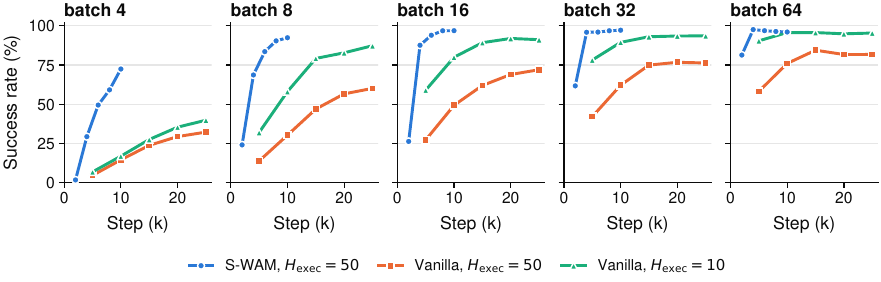}
\caption{Success rate versus training steps across five global batch sizes.}
\label{fig:budget}
\end{figure}

\paragraph{Training efficiency.}
\label{sec:exp:budget}
Training under \ours{}'s streaming schedule is more expensive per step, but converges substantially faster. Across five global batch sizes, \ourmodel{} consistently reaches higher success with fewer training steps, with the largest gains in low-batch regimes (Fig.~\ref{fig:budget}). At batch size $4$, it reaches $72.5$ success after $10$k steps, compared with $32.2$ for vanilla even after $25$k steps; the gap narrows to $14.3$ points at batch size $64$. Although each streaming-training step costs $1.87\times$ more FLOPs (Appendix~\ref{app:setup-train}), the shorter schedule more than offsets this overhead: \ourmodel{} ends at a higher success rate than vanilla while spending fewer total training FLOPs (Table~\ref{tab:app-train-budget}).

\begin{figure*}[t]
\centering

\begin{minipage}[t]{0.59\textwidth}
\vspace{0pt}
\centering

\captionof{table}{Optimization ladder on the LaWAM host.}
\label{tab:speed}
\resizebox{\linewidth}{!}{%
\begin{tabular}{lrrr}
\toprule
Method & Action/s $\uparrow$ & Speedup & TTFA $\downarrow$ \\
\midrule
Vanilla ($H_{\text{exec}}{=}10$), eager & 80.9\,$\pm$\,1.0 & 1.00$\times$ & 123.6 \\
Vanilla ($H_{\text{exec}}{=}50$), eager & 402.5\,$\pm$\,6.8 & 4.98$\times$ & 124.2 \\
\textbf{Ours} ($H_{\text{exec}}{=}50$), eager & 292.7\,$\pm$\,9.4 & 3.62$\times$ & 73.3 \\
\quad $+$ geo \& DiT CUDA Graph & 625.7\,$\pm$\,16.1 & 7.73$\times$ & 53.2 \\
\quad $+$ prompt cache & 642.9\,$\pm$\,18.2 & 7.95$\times$ & 48.6 \\
\bottomrule
\end{tabular}%
}

\end{minipage}
\hfill
\begin{minipage}[t]{0.39\textwidth}
\vspace{-8pt}
\centering

\includegraphics[width=\linewidth]{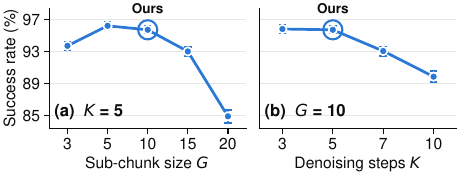}

\captionof{figure}{Effect of chunking hyper-parameters, where $H = K \times G$.}
\label{fig:kgran}
\end{minipage}

\vspace{-0.2cm}
\end{figure*}

\paragraph{Inference speed.} Against the vanilla policy at the execution horizon that matches its accuracy ($H_{\mathrm{exec}}{=}10$), \ourmodel{} raises throughput from $80.9$ to $292.7$ executed actions per second ($3.62\times$) and cuts TTFA from $123.6$ to $73.3$\,ms ($-40.7\%$). Adding graph capture and prompt caching, neither of which changes the actions produced, takes the same policy to $642.9$ actions per second at $48.6$\,ms (Table~\ref{tab:speed}). 
Here, \textit{eager} uses standard PyTorch execution, \textit{geo \& DiT CUDA Graph} captures the geometry and denoising calls to reduce launch overhead, and \textit{prompt cache} reuses the frame-invariant prompt computation. The measurement protocol is provided in Appendix~\ref{app:setup-speed}.

\section{Discussion}
\label{sec:discussion}

\subsection{Execution Length and the Speed--Accuracy Trade-off}
\label{sec:disc:execlen}
\begin{wrapfigure}{r}{0.38\textwidth}
\vspace{-0.2cm}
\centering
\includegraphics[width=0.38\textwidth]{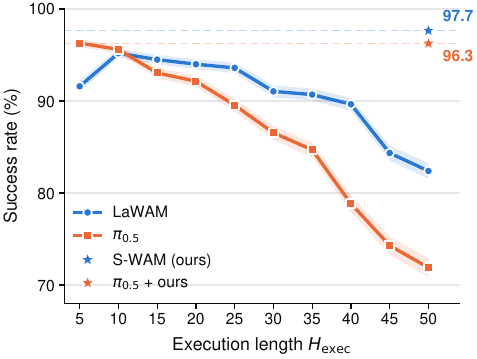}
\vspace{-0.6cm}
\caption{Success rate versus execution horizon, averaged over four suites.}
\label{fig:exec}
\vspace{-0.2cm}
\end{wrapfigure}
Increasing throughput by executing more actions per inference substantially degrades conventional policies. With only $H_{\mathrm{exec}}$ varied, our backbone drops from $95.2$ at $H_{\mathrm{exec}}{=}10$ to $82.4$ at $50$, while $\pi_{0.5}$ drops from $93.2$ to $66.4$ (Table~\ref{tab:hosts}). In contrast, \ourmodel{} executes the full $50$-action chunk and still reaches $97.7$ and $96.5$ on the two backbones, above the best vanilla result on either. Thus, the throughput improvement does not come from sacrificing task performance. 
This suggests that the main limitation lies in executing later actions under stale conditioning, rather than in the chunk length itself. \ourmodel{} mitigates this by refreshing the conditioning for unexecuted actions as new observations arrive. 
Vanilla SR peaks at or near $H_{\mathrm{exec}}{=}10$ on both backbones (Fig.~\ref{fig:exec}), so we compare mainly against that configuration.
Details are in Appendix~\ref{app:exec}.

\subsection{Choice of Denoising Steps and Sub-Chunk Size}
\label{sec:disc:kgran}

The total chunk length is determined by the number of denoising steps $K$ and the sub-chunk size $G$, the number of actions emitted per step. Increasing either increases the executed horizon and thus throughput. We vary each factor independently while keeping all other settings fixed. As shown in Fig.~\ref{fig:kgran}, performance remains stable over a broad range and degrades only when the chunk becomes too long relative to the available denoising steps. We therefore use $K{=}5$ and $G{=}10$, yielding a $50$-action chunk, for all experiments unless otherwise specified.


\begin{figure*}[htbp]
    \centering
    \includegraphics[width=\textwidth]{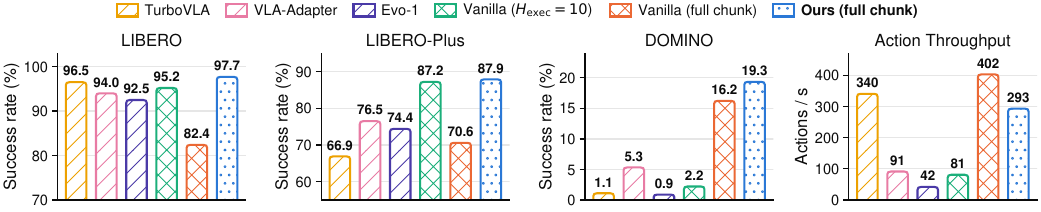}
    \caption{Comparison with compact VLA models on LIBERO, LIBERO-Plus, DOMINO.}
    \label{fig:small-model-compare}
\end{figure*}

\subsection{Small Models vs.\ Execution-Level Acceleration}

We compare with three compact policies, TurboVLA~\citep{turbovla2026} ($0.2$B), VLA-Adapter~\citep{vlaadapter2025} ($0.6$B), and Evo-1~\citep{evo1_2025} ($0.77$B), against our $2.3$B backbone. On LIBERO, the compact models already perform strongly (Fig.~\ref{fig:small-model-compare}). The gap widens on harder benchmarks: the compact models achieve 66.9--76.5\% on LIBERO-Plus versus 87.9\% for ours, and 0.9--5.3\% on DOMINO, where \ourmodel{} reaches 19.3\%. 
Thus, these results highlight two complementary paths to efficient policy inference: reducing model size, which works well on relatively easier settings such as LIBERO~\citep{wang2026visionlanguageaction}, and execution-level acceleration, which preserves the capacity of a larger policy. By retaining a larger policy while increasing the number of reliable actions executed per expensive policy inference, \ourmodel{} improves throughput while preserving stronger performance on challenging robustness and dynamic benchmarks. Detailed results are provided in Appendix~\ref{app:detailed-results}.

\subsection{The Future-Prediction Error as a Chunking Signal}
\label{sec:disc:signal}
\begin{wrapfigure}{r}{0.36\textwidth}
\vspace{-0.6cm}
\centering
\includegraphics[width=0.36\textwidth]{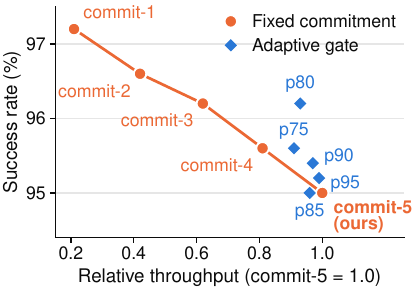}
\vspace{-0.55cm}
\caption{SR against throughput for fixed and adaptive commitment on LIBERO-Long.}
\label{fig:gate}
\vspace{-0.4cm}
\end{wrapfigure}

We test whether the future-prediction error $\delta_j$ can guide adaptive chunking. \emph{Commit-$X$} always executes exactly $X$ sub-chunks before replanning. \emph{p$X$} instead sets the gating threshold to the $X$-th percentile of $\delta_j$ measured on the validation set, and replans when the error exceeds this threshold, while committing to at least four sub-chunks. 
As shown in Fig.~\ref{fig:gate}, adaptive chunking provides an additional operating point between commit-4 and commit-5, trading a small amount of throughput for a larger gain in SR. For example, p80 reaches $96.2$ SR with only a modest further reduction in throughput. 
These results suggest that future-prediction error can serve as a practical signal for adaptive chunking.

\subsection{Transferability Across Policy Backbones}
\label{sec:disc:hosts}

We apply \ours{} to three different VLA backbones, LaWAM~\citep{chen2026lawam}, $\pi_{0.5}$~\citep{intelligence2025pi05}, and FLOWER~\citep{reuss2025flower}. Each backbone retains its host-specific training configuration, while the \ours{} schedule, objectives, and method-side hyper-parameters remain unchanged. Training details are provided in Appendix~\ref{app:setup-train}. Compared with vanilla execution at $H_{\mathrm{exec}}{=}50$, \ourmodel{} improves SR by $15.25$, $30.15$, and $14.05$ points on LaWAM, $\pi_{0.5}$, and FLOWER, respectively (Table~\ref{tab:hosts}). Notably, $\pi_{0.5}$ and FLOWER were not pretrained to predict future latents, suggesting that the method does not rely on such pretraining. Across all three backbones, \ourmodel{} achieves comparable or better SR than vanilla execution at $H_{\mathrm{exec}}{=}10$, while improving throughput by $2.74\times$--$3.62\times$ and reducing TTFA by $37.9\%$--$40.7\%$.

\begin{table}[htbp]
\centering\small
\caption{\ours{} transfers across different backbones. Act/s is executed actions per second; TTFA is time to first action (ms). }
\label{tab:hosts}
\setlength{\tabcolsep}{4pt}
\begin{tabular}{llcccccrr}
\toprule
& & \multicolumn{5}{c}{Success rate} & \multicolumn{2}{c}{Speed} \\
\cmidrule(lr){3-7}\cmidrule(lr){8-9}
Backbone & Policy & Spatial & Object & Goal & Long & Avg. & Act/s\,$\uparrow$ & TTFA$\downarrow$ \\
\midrule
LaWAM & \ourmodel & \textbf{98.6} & \textbf{100.0} & \textbf{97.0} & \textbf{95.0} & \textbf{97.65} & 292.7 & \textbf{73.3} \\
\citep{chen2026lawam} & Vanilla, $H_{\mathrm{exec}}{=}10$ & 96.2 & 98.4 & 95.6 & 90.6 & 95.20 & 80.9 & 123.6 \\
(2.3B) & Vanilla, $H_{\mathrm{exec}}{=}50$ & 87.2 & 80.6 & 88.0 & 73.8 & 82.40 & \textbf{402.5} & 124.2 \\
\midrule
$\pi_{0.5}$ & \ourmodel & \textbf{98.0} & 98.6 & \textbf{95.2} & \textbf{94.2} & \textbf{96.50} & 243.2 & \textbf{80.6} \\
\citep{intelligence2025pi05} & Vanilla, $H_{\mathrm{exec}}{=}10$ & 94.6 & \textbf{99.0} & 91.6 & 87.4 & 93.15 & 75.7 & 132.2 \\
(3.5B) & Vanilla, $H_{\mathrm{exec}}{=}50$ & 66.4 & 67.6 & 72.8 & 58.6 & 66.35 & \textbf{378.3} & 134.8 \\
\midrule
FLOWER & \ourmodel & 97.4 & \textbf{99.6} & \textbf{98.6} & 88.8 & \textbf{96.10} & 234.3 & \textbf{72.7} \\
\citep{reuss2025flower} & Vanilla, $H_{\mathrm{exec}}{=}10$ & \textbf{98.8} & 99.2 & 96.6 & \textbf{89.6} & 96.05 & 85.5 & 117.0 \\
(1.0B) & Vanilla, $H_{\mathrm{exec}}{=}50$ & 85.6 & 85.0 & 90.6 & 67.0 & 82.05 & \textbf{416.0} & 120.2 \\
\bottomrule
\end{tabular}

\end{table}

\section{Conclusion}
\label{sec:conclusions}

We presented \ours, a streaming inference and training framework for World-Action Models that pipelines the generation and execution of large action chunks. \ours{} maintains sub-chunks at staggered denoising stages and continuously updates unexecuted actions using new observations, enabling longer execution horizons without repeatedly invoking the full policy. \ourmodel{} achieves strong performance across multiple simulation benchmarks and real-robot tasks while substantially improving action throughput and reducing TTFA without sacrificing task performance. It also improves robustness under perturbations and consistently outperforms compute-matched baselines. Overall, these results show that streaming generation and observation-conditioned refinement provide an effective way to make large action chunks practical for efficient robot control.

\clearpage
\subsection*{AI Use Statement}

Generative AI tools were used in this work in three ways. First, for language polishing: a large language model was used to improve the clarity and grammar of the manuscript, and every suggested edit was reviewed and accepted or rejected by the authors. Second, for experiment management: AI assistance was used to launch, schedule and monitor training and evaluation runs across machines. Third, for result aggregation: AI assistance was used to collect per-run outputs into the tables and figures reported here. Every number obtained in this way was manually checked by the authors against the raw evaluation logs before being reported. The method, its implementation, the experimental design and all scientific claims are the authors' own work, and the authors take full responsibility for the content of this paper.

\subsection*{Reproducibility Statement}

The method is specified in Sec.~\ref{sec:method}, with the staircase schedule, the refresh step and the training objective given in Sec.~\ref{sec:method-staircase}--\ref{sec:method-training} and Algorithm~\ref{alg:streamwam}. The appendix records what is needed to reproduce every reported number. Per-backbone training recipes, including the optimiser, the schedule, the step budgets and the boundary grid, are in Appendix~\ref{app:setup-train}; the latency and throughput measurement protocol is in Appendix~\ref{app:setup-speed}. Appendix~\ref{app:libero} gives the full per-method LIBERO table behind Fig.~\ref{fig:teaser}, including the entries omitted from the figure, and Appendix~\ref{app:exec} the per-suite execution-length curves. Per-task results for every benchmark, together with the compact-model baselines, are in Appendix~\ref{app:detailed-results}. The real-robot datasets, the observation and action spaces, the scene-randomisation and success criteria, and the per-task success rates are in Appendix~\ref{app:real-datasets}.

\subsection*{Ethics Statement}

This work involves no human subjects, no personally identifying data and no newly released dataset. 

\bibliography{refs}

@article{chen2026lawam,
  title   = {{LaWAM: Latent World Action Models for Efficient Dynamics-Aware Robot Policies}},
  author  = {Jialei Chen and Kai Wang and Kang Chen and Shuaihang Chen and Feng Gao and Wenhao Tang and Zhiyuan Li and Weilin Liu and Zhuyu Yao and Boxun Li and Yuanbo Xu and Chao Yu},
  journal = {arXiv preprint arXiv:2606.15768},
  year    = {2026}
}

@article{liu2023libero,
  title   = {{LIBERO: Benchmarking Knowledge Transfer for Lifelong Robot Learning}},
  author  = {Bo Liu and Yifeng Zhu and Chongkai Gao and Yihao Feng and Qiang Liu and Yuke Zhu and Peter Stone},
  journal = {arXiv preprint arXiv:2306.03310},
  year    = {2023},
  note    = {NeurIPS Datasets and Benchmarks 2023}
}

@article{black2024pi0,
  title   = {{$\pi_0$: A Vision-Language-Action Flow Model for General Robot Control}},
  author  = {Kevin Black and Noah Brown and Danny Driess and Adnan Esmail and Michael Equi and Chelsea Finn and Niccolo Fusai and Lachy Groom and Karol Hausman and Brian Ichter and others},
  journal = {arXiv preprint arXiv:2410.24164},
  year    = {2024},
  note    = {RSS 2025}
}

@article{kim2024openvla,
  title   = {{OpenVLA: An Open-Source Vision-Language-Action Model}},
  author  = {Moo Jin Kim and Karl Pertsch and Siddharth Karamcheti and Ted Xiao and Ashwin Balakrishna and Suraj Nair and Rafael Rafailov and Ethan Foster and Grace Lam and Pannag Sanketi and others},
  journal = {arXiv preprint arXiv:2406.09246},
  year    = {2024}
}

@article{shukor2025smolvla,
  title   = {{SmolVLA: A Vision-Language-Action Model for Affordable and Efficient Robotics}},
  author  = {Mustafa Shukor and Dana Aubakirova and Francesco Capuano and Pepijn Kooijmans and Steven Palma and Adil Zouitine and Michel Aractingi and Caroline Pascal and Martino Russi and Andres Marafioti and others},
  journal = {arXiv preprint arXiv:2506.01844},
  year    = {2025}
}

@article{wang2026ffdc,
  title   = {{When to Trust Imagination: Adaptive Action Execution for World Action Models}},
  author  = {Rui Wang and Yue Zhang and Jiehong Lin and Kuncheng Luo and Jianan Wang and Zhongrui Wang and Xiaojuan Qi},
  journal = {arXiv preprint arXiv:2605.06222},
  year    = {2026}
}

@article{yuan2026fastwam,
  title   = {{Fast-WAM: Do World Action Models Need Test-time Future Imagination?}},
  author  = {Tianyuan Yuan and Zibin Dong and Yicheng Liu and Hang Zhao},
  journal = {arXiv preprint arXiv:2603.16666},
  year    = {2026}
}

@article{hoeg2024sdp,
  title   = {{Streaming Diffusion Policy: Fast Policy Synthesis with Variable Noise Diffusion Models}},
  author  = {Sigmund H. Høeg and Yilun Du and Olav Egeland},
  journal = {arXiv preprint arXiv:2406.04806},
  year    = {2025},
  note    = {ICRA 2025}
}

@article{chen2024diffusionforcing,
  title   = {{Diffusion Forcing: Next-token Prediction Meets Full-Sequence Diffusion}},
  author  = {Boyuan Chen and Diego Marti Monso and Yilun Du and Max Simchowitz and Russ Tedrake and Vincent Sitzmann},
  journal = {arXiv preprint arXiv:2407.01392},
  year    = {2024},
  note    = {NeurIPS 2024}
}

@article{chen2025rnrdp,
  title   = {{Responsive Noise-Relaying Diffusion Policy: Responsive and Efficient Visuomotor Control}},
  author  = {Zhuoqun Chen and Xiu Yuan and Tongzhou Mu and Hao Su},
  journal = {arXiv preprint arXiv:2502.12724},
  year    = {2025}
}

@article{shi2026streamingvla,
  title   = {{StreamingVLA: Streaming Vision-Language-Action Model with Action Flow Matching and Adaptive Early Observation}},
  author  = {Yiran Shi and Dongqi Guo and Tianchen Zhao and Feng Gao and Liangzhi Shi and Chao Yu and ZhiJian Mo and Qihua Xiao and XiaoShuai Peng and Qingmin Liao and Yu Wang},
  journal = {arXiv preprint arXiv:2603.28565},
  year    = {2026}
}

@article{song2026faster,
  title   = {{FASTER: Rethinking Real-Time Flow VLAs}},
  author  = {Yuxiang Lu and Zhe Liu and Xianzhe Fan and Zhenya Yang and Jinghua Hou and Junyi Li and Kaixin Ding and Hengshuang Zhao},
  journal = {arXiv preprint arXiv:2603.19199},
  year    = {2026}
}

@article{black2025rtc,
  title   = {{Real-Time Execution of Action Chunking Flow Policies}},
  author  = {Kevin Black and Manuel Y. Galliker and Sergey Levine},
  journal = {arXiv preprint arXiv:2506.07339},
  year    = {2025},
  note    = {NeurIPS 2025}
}

@article{black2025ttrtc,
  title   = {{Training-Time Action Conditioning for Efficient Real-Time Chunking}},
  author  = {Kevin Black and Allen Z. Ren and Michael Equi and Sergey Levine},
  journal = {arXiv preprint arXiv:2512.05964},
  year    = {2025}
}

@article{park2026pir2,
  title   = {{$\pi\mathbf{R}^2$: Reactive Real-time Flow Policies}},
  author  = {Sungjae Park and Shubham Tulsiani},
  journal = {arXiv preprint arXiv:2607.26055},
  year    = {2026}
}

@article{liu2024bid,
  title   = {{Bidirectional Decoding: Improving Action Chunking via Guided Test-Time Sampling}},
  author  = {Yuejiang Liu and Jubayer Ibn Hamid and Annie Xie and Yoonho Lee and Maximilian Du and Chelsea Finn},
  journal = {arXiv preprint arXiv:2408.17355},
  year    = {2024},
  note    = {ICLR 2025}
}

@article{sendai2025a2c2,
  title   = {{Leave No Observation Behind: Real-time Correction for VLA Action Chunks}},
  author  = {Kohei Sendai and Maxime Alvarez and Tatsuya Matsushima and Yutaka Matsuo and Yusuke Iwasawa},
  journal = {arXiv preprint arXiv:2509.23224},
  year    = {2025}
}

@article{pan2026vlacorrector,
  title   = {{VLA-Corrector: Lightweight Detect-and-Correct Inference for Adaptive Action Horizon}},
  author  = {Yi Pan and Miao Pan and Qi Lu and Jiaming Huang and Man Zhang and Siteng Huang and Xin Li and Jie Zhang and Yongliang Shen and Xuhong Zhang and Wenqi Zhang},
  journal = {arXiv preprint arXiv:2607.01804},
  year    = {2026}
}

@article{yang2026jetsonpi,
  title   = {{Jetson-PI: Towards Onboard Real-Time Robot Control via Foresight-Aligned Asynchronous Inference}},
  author  = {Zebin Yang and Qi Wang and Yunhe Wang and Xiurui Guo and Bo Yu and Shaoshan Liu and Jiafeng Xu and Hao Dong and Meng Li},
  journal = {arXiv preprint arXiv:2607.12659},
  year    = {2026}
}

@article{feng2026dvac,
  title   = {{Denoising Tells When to Replan: Denoising-Variance Adaptive Chunking for Flow-Based Robot Policies}},
  author  = {Xiangdong Feng and Yuxuan Cheng and Chen Shi and Boyao Han and Yuxuan Yan and Yitong Hong and Zhuotao Tian and Li Jiang},
  journal = {arXiv preprint arXiv:2606.03847},
  year    = {2026}
}

@article{kim2025openvlaoft,
  title   = {{Fine-Tuning Vision-Language-Action Models: Optimizing Speed and Success}},
  author  = {Moo Jin Kim and Chelsea Finn and Percy Liang},
  journal = {arXiv preprint arXiv:2502.19645},
  year    = {2025},
  note    = {RSS 2025}
}

@article{luan2026snapflow,
  title   = {{SnapFlow: One-Step Action Generation for Flow-Matching VLAs via Progressive Self-Distillation}},
  author  = {Wuyang Luan and Junhui Li and Weiguang Zhao and Wenjian Zhang and Tieru Wu and Rui Ma},
  journal = {arXiv preprint arXiv:2604.05656},
  year    = {2026}
}

@article{chen2025fisvla,
  title   = {{Fast-in-Slow: A Dual-System Foundation Model Unifying Fast Manipulation within Slow Reasoning}},
  author  = {Hao Chen and Jiaming Liu and Chenyang Gu and Zhuoyang Liu and Renrui Zhang and Xiaoqi Li and Xiao He and Yandong Guo and Chi-Wing Fu and Shanghang Zhang and Pheng-Ann Heng},
  journal = {arXiv preprint arXiv:2506.01953},
  year    = {2025}
}

@article{
wang2026visionlanguageaction,
title={Vision-Language-Action in Robotics: A Survey of Datasets, Benchmarks, and Data Engines},
author={Ziyao Wang and Bingying Wang and Hanrong Zhang and Tingting Du and Tianyang Chen and Guoheng Sun and Yexiao He and Zheyu Shen and Wanghao Ye and Ang Li},
journal={Transactions on Machine Learning Research},
issn={2835-8856},
year={2026},
url={https://openreview.net/forum?id=tAaWFpvnmm},
note={Featured Certification, Reproducibility Certification, Survey Certification}
}

@article{zhao2023act,
  title   = {{Learning Fine-Grained Bimanual Manipulation with Low-Cost Hardware}},
  author  = {Tony Z. Zhao and Vikash Kumar and Sergey Levine and Chelsea Finn},
  journal = {arXiv preprint arXiv:2304.13705},
  year    = {2023},
  note    = {RSS 2023}
}

@article{chi2023diffusionpolicy,
  title   = {{Diffusion Policy: Visuomotor Policy Learning via Action Diffusion}},
  author  = {Cheng Chi and Zhenjia Xu and Siyuan Feng and Eric Cousineau and Yilun Du and Benjamin Burchfiel and Russ Tedrake and Shuran Song},
  journal = {arXiv preprint arXiv:2303.04137},
  year    = {2023},
  note    = {RSS 2023}
}

@article{lipman2023flowmatching,
  title   = {{Flow Matching for Generative Modeling}},
  author  = {Yaron Lipman and Ricky T. Q. Chen and Heli Ben-Hamu and Maximilian Nickel and Matt Le},
  journal = {arXiv preprint arXiv:2210.02747},
  year    = {2022},
  note    = {ICLR 2023}
}

@article{liu2023rectifiedflow,
  title   = {{Flow Straight and Fast: Learning to Generate and Transfer Data with Rectified Flow}},
  author  = {Xingchao Liu and Chengyue Gong and Qiang Liu},
  journal = {arXiv preprint arXiv:2209.03003},
  year    = {2022},
  note    = {ICLR 2023}
}

@article{intelligence2025pi05,
  title   = {{$\pi_{0.5}$: a Vision-Language-Action Model with Open-World Generalization}},
  author  = {Physical Intelligence and Kevin Black and Noah Brown and James Darpinian and Karan Dhabalia and Danny Driess and Adnan Esmail and Michael Equi and Chelsea Finn and Niccolo Fusai and others},
  journal = {arXiv preprint arXiv:2504.16054},
  year    = {2025}
}

@article{du2023unipi,
  title   = {{Learning Universal Policies via Text-Guided Video Generation}},
  author  = {Yilun Du and Mengjiao Yang and Bo Dai and Hanjun Dai and Ofir Nachum and Joshua B. Tenenbaum and Dale Schuurmans and Pieter Abbeel},
  journal = {arXiv preprint arXiv:2302.00111},
  year    = {2023},
  note    = {NeurIPS 2023}
}

@article{bruce2024genie,
  title   = {{Genie: Generative Interactive Environments}},
  author  = {Jake Bruce and Michael Dennis and Ashley Edwards and Jack Parker-Holder and Yuge Shi and Edward Hughes and Matthew Lai and Aditi Mavalankar and Richie Steigerwald and Chris Apps and others},
  journal = {arXiv preprint arXiv:2402.15391},
  year    = {2024},
  note    = {ICML 2024}
}

@article{ahead2026,
  title   = {{Intercepting the Future: Latent-Space Predictive World Model for Dynamic VLA Manipulation}},
  author  = {Shahram Najam Syed and Arthur Jakobsson and Haoran Hao and Jeffrey Ichnowski},
  journal = {arXiv preprint arXiv:2606.02486},
  year    = {2026}
}

@article{streamvla2026,
  title   = {{StreamVLA: Breaking the Reason-Act Cycle via Completion-State Gating}},
  author  = {Tongqing Chen and Hang Wu and Jiasen Wang and Xiaotao Li and Lu Fang},
  journal = {arXiv preprint arXiv:2602.01100},
  year    = {2026}
}

@article{motubrain2026,
  title   = {{Motubrain: An Advanced World Action Model for Robot Control}},
  author  = { Motubrain Team and Chendong Xiang and Fan Bao and Haitian Liu and Hengkai Tan and Hongzhe Bi and James Li and Jiabao Liu and Jingrui Pang and Kiro Jing and others},
  journal = {arXiv preprint arXiv:2604.27792},
  year    = {2026}
}

@article{ddp2026,
  title   = {{Dreaming the Unseen: World Model-regularized Diffusion Policy for Out-of-Distribution Robustness}},
  author  = {Ziou Hu and Xiangtong Yao and Yuan Meng and Zhenshan Bing and Alois Knoll},
  journal = {arXiv preprint arXiv:2603.21017},
  year    = {2026}
}

@article{pearlvla2026,
  title   = {{PearlVLA: Progressive Embodied Action-Plan Refinement in Latent Space}},
  author  = {Bochen Yang and Lianlei Shan},
  journal = {arXiv preprint arXiv:2606.17924},
  year    = {2026}
}

@article{jepavla2026,
  title   = {{JEPA-VLA: Video Predictive Embedding is Needed for VLA Models}},
  author  = {Shangchen Miao and Ningya Feng and Jialong Wu and Ye Lin and Xu He and Dong Li and Mingsheng Long},
  journal = {arXiv preprint arXiv:2602.11832},
  year    = {2026}
}

@article{flare2025,
  title   = {{FLARE: Robot Learning with Implicit World Modeling}},
  author  = {Ruijie Zheng and Jing Wang and Scott Reed and Johan Bjorck and Yu Fang and Fengyuan Hu and Joel Jang and Kaushil Kundalia and Zongyu Lin and Loic Magne and others},
  journal = {arXiv preprint arXiv:2505.15659},
  year    = {2025}
}

@article{wamrobust2026,
  title   = {{Do World Action Models Generalize Better than VLAs? A Robustness Study}},
  author  = {Zhanguang Zhang and Zhiyuan Li and Behnam Rahmati and Rui Heng Yang and Yintao Ma and Amir Rasouli and Sajjad Pakdamansavoji and Yangzheng Wu and Lingfeng Zhang and Tongtong Cao and others},
  journal = {arXiv preprint arXiv:2603.22078},
  year    = {2026}
}

@article{asyncbench2026,
  title   = {{Understanding Asynchronous Inference Methods for Vision-Language-Action Models}},
  author  = {Ayoub Agouzoul},
  journal = {arXiv preprint arXiv:2605.08168},
  year    = {2026}
}

@article{sants2026,
  title   = {{SANTS: A State-Adaptive Scheduler for World Action Models}},
  author  = {Yirui Sun and Guangyu Zhuge and Keliang Liu and Jie Gu and Shiqin Dai and Xinyu Bing and Zhongxue Gan and Chunxu Tian},
  journal = {arXiv preprint arXiv:2605.27947},
  year    = {2026}
}

@article{liberoplus2025,
  title   = {{LIBERO-Plus: In-depth Robustness Analysis of Vision-Language-Action Models}},
  author  = {Senyu Fei and Siyin Wang and Junhao Shi and Zihao Dai and Jikun Cai and Pengfang Qian and Li Ji and Xinzhe He and Shiduo Zhang and Zhaoye Fei and others},
  journal = {arXiv preprint arXiv:2510.13626},
  year    = {2025}
}

@article{noisegate2026,
  title   = {{NoiseGate: Learning Per-Latent Timestep Schedules as Information Gating in World Action Models}},
  author  = {Wen Huang and Haoran Sun and Yongjian Guo and Yunxuan Ma and Haoran Li and Jing Long and Zhouying Mo and Zhong Guan and Yucheng Guo and Shuai Di and Junwu Xiong},
  journal = {arXiv preprint arXiv:2605.07794},
  year    = {2026}
}

@article{remac2026,
  title   = {{Real-Time Robot Execution with Masked Action Chunking}},
  author  = {Haoxuan Wang and Gengyu Zhang and Yan Yan and Yuzhang Shang and Ramana Rao Kompella and Gaowen Liu},
  journal = {arXiv preprint arXiv:2601.20130},
  year    = {2026}
}

@article{tidal2026,
  title   = {{TIDAL: Temporally Interleaved Diffusion and Action Loop for High-Frequency VLA Control}},
  author  = {Yuteng Sun and Haoran Wang and Ruofei Bai and Zhengguo Li and Jun Li and Meng Yee Michael Chuah and Wei Yun Yau},
  journal = {arXiv preprint arXiv:2601.14945},
  year    = {2026}
}

@article{wang2026stwam,
  title   = {{ST-WAM: Semantic-Temporal World Action Model for Robust Manipulation under Visual Distribution Shifts}},
  author  = {Mingxin Wang and Bin Hu and Bin Qian and Kaitao Jiang and Haoning Wu and Feng Yan and Bowen Jing and Ruiyang Hao and Enyi Wang and Kangning Niu and others},
  journal = {arXiv preprint arXiv:2607.28993},
  year    = {2026}
}

@misc{sun2026dropthenrecovery,
      title={Drop-Then-Recovery: How Redundant Are Vision-Language-Action Models?}, 
      author={Guoheng Sun and Kaixi Feng and Shwai He and Xiaochuan Gong and Yexiao He and Ziyao Wang and Zheyu Shen and Wanghao Ye and Ramana Rao Kompella and Gaowen Liu and Ang Li},
      year={2026},
      eprint={2606.27755},
      archivePrefix={arXiv},
      primaryClass={cs.RO},
      url={https://arxiv.org/abs/2606.27755}, 
}

@article{yuan2026dreamwam,
  title   = {{DreamWAM: Beyond RGB Future Prediction for World Action Models}},
  author  = {Shanglin Yuan and Weiheng Zhao and Xin Shi and Haoyi Jiang and Xianda Guo and Liu Liu and Wenyu Liu and Wei Sui and Xinggang Wang},
  journal = {arXiv preprint arXiv:2608.04996},
  year    = {2026}
}

@article{sun2026vlajepa,
  title   = {{VLA-JEPA: Enhancing Vision-Language-Action Model with Latent World Model}},
  author  = {Jingwen Sun and Wenyao Zhang and Zekun Qi and Shaojie Ren and Zezhi Liu and Hanxin Zhu and Guangzhong Sun and Xin Jin and Zhibo Chen},
  journal = {arXiv preprint arXiv:2602.10098},
  year    = {2026}
}

@article{sun2026rocket,
  title   = {{ROCKET: Residual-Oriented Multi-Layer Alignment for Spatially-Aware Vision-Language-Action Models}},
  author  = {Guoheng Sun and Tingting Du and Kaixi Feng and Chenxiang Luo and Xingguo Ding and Zheyu Shen and Ziyao Wang and Yexiao He and Ang Li},
  journal = {arXiv preprint arXiv:2602.17951},
  year    = {2026}
}

@article{wu2026vlanext,
  title   = {{VLANeXt: Recipes for Building Strong VLA Models}},
  author  = {Xiao-Ming Wu and Bin Fan and Kang Liao and Jian-Jian Jiang and Runze Yang and Yihang Luo and Zhonghua Wu and Wei-Shi Zheng and Chen Change Loy},
  journal = {arXiv preprint arXiv:2602.18532},
  year    = {2026},
  note    = {ICML 2026}
}

@article{lin2026worldpilot,
  title   = {{World Pilot: Steering Vision-Language-Action Models with World-Action Priors}},
  author  = {Zefu Lin and Rongxu Cui and Junjia Xu and Xiaojuan Jin and Wenling Li and Lue Fan and Zhaoxiang Zhang},
  journal = {arXiv preprint arXiv:2606.12403},
  year    = {2026}
}

@article{ma2026internvlaa15,
  title   = {{InternVLA-A1.5: Unifying Understanding, Latent Foresight, and Action for Compositional Generalization}},
  author  = {Haoxiang Ma and Junhao Cai and Xiaoxu Xu and Hao Li and Yuyin Yang and Yang Tian and Jiafei Cao and Hongrui Zhu and Zherui Qiu and Zhaxizhuoma and others},
  journal = {arXiv preprint arXiv:2607.04988},
  year    = {2026}
}

@article{chen2025internvlam1,
  title   = {{InternVLA-M1: A Spatially Guided Vision-Language-Action Framework for Generalist Robot Policy}},
  author  = {Xinyi Chen and Yilun Chen and Yanwei Fu and Ning Gao and Jiaya Jia and Weiyang Jin and Hao Li and Yao Mu and Jiangmiao Pang and Yu Qiao and others},
  journal = {arXiv preprint arXiv:2510.13778},
  year    = {2025}
}

@article{lin2026jepawam,
  title   = {{JEPA-WAM: Learning Vision-Language-Action Policies with Joint-Embedding World Modeling}},
  author  = {Yihan Lin and Jiawei He and Shifeng Bao and Chen Zhao and Yang Li and Xiaobo Wang and Yan Wang and Cheng Chi and Jing Zhang},
  journal = {arXiv preprint arXiv:2608.09381},
  year    = {2026}
}

@article{lin2026evodepth,
  title   = {{Evo-Depth: A Lightweight Depth-Enhanced Vision-Language-Action Model}},
  author  = {Tao Lin and Yuxin Du and Jiting Liu and Nuobei Zhu and Yunhe Li and Yuqian Fu and Yinxinyu Chen and Hongyi Cai and Zewei Ye and Bing Cheng and others},
  journal = {arXiv preprint arXiv:2605.14950},
  year    = {2026}
}

@article{reuss2025flower,
  title   = {{FLOWER: Democratizing Generalist Robot Policies with Efficient Vision-Language-Action Flow Policies}},
  author  = {Moritz Reuss and Hongyi Zhou and Marcel Rühle and Ömer Erdinç Yağmurlu and Fabian Otto and Rudolf Lioutikov},
  journal = {arXiv preprint arXiv:2509.04996},
  year    = {2025},
  note    = {CoRL 2025}
}

@article{wang2025bitvla,
  title   = {{BitVLA: 1-bit Vision-Language-Action Models for Robotics Manipulation}},
  author  = {Hongyu Wang and Chuyan Xiong and Ruiping Wang and Xilin Chen},
  journal = {arXiv preprint arXiv:2506.07530},
  year    = {2025}
}

@article{bu2025univla,
  title   = {{UniVLA: Learning to Act Anywhere with Task-centric Latent Actions}},
  author  = {Qingwen Bu and Yanting Yang and Jisong Cai and Shenyuan Gao and Guanghui Ren and Maoqing Yao and Ping Luo and Hongyang Li},
  journal = {arXiv preprint arXiv:2505.06111},
  year    = {2025},
  note    = {RSS 2025}
}

@article{bi2025motus,
  title   = {{Motus: A Unified Latent Action World Model}},
  author  = {Hongzhe Bi and Hengkai Tan and Shenghao Xie and Zeyuan Wang and Shuhe Huang and Haitian Liu and Ruowen Zhao and Yao Feng and Chendong Xiang and Yinze Rong and others},
  journal = {arXiv preprint arXiv:2512.13030},
  year    = {2025}
}

@article{hung2025nora,
  title   = {{NORA: A Small Open-Sourced Generalist Vision Language Action Model for Embodied Tasks}},
  author  = {Chia-Yu Hung and Qi Sun and Pengfei Hong and Amir Zadeh and Chuan Li and U-Xuan Tan and Navonil Majumder and Soujanya Poria},
  journal = {arXiv preprint arXiv:2504.19854},
  year    = {2025}
}

@article{qu2025spatialvla,
  title   = {{SpatialVLA: Exploring Spatial Representations for Visual-Language-Action Model}},
  author  = {Delin Qu and Haoming Song and Qizhi Chen and Yuanqi Yao and Xinyi Ye and Yan Ding and Zhigang Wang and JiaYuan Gu and Bin Zhao and Dong Wang and Xuelong Li},
  journal = {arXiv preprint arXiv:2501.15830},
  year    = {2025}
}

@article{zhang2026a1,
  title   = {{A1: A Fully Transparent Open-Source, Adaptive and Efficient Truncated Vision-Language-Action Model}},
  author  = {Kaidong Zhang and Jian Zhang and Rongtao Xu and Yu Sun and Shuoshuo Xue and Youpeng Wen and Xiaoyu Guo and Minghao Guo and Weijia Liufu and Liu Zihou and others},
  journal = {arXiv preprint arXiv:2604.05672},
  year    = {2026}
}

@misc{minicpm2026robotmanip,
  title        = {{MiniCPM-RobotManip}},
  author       = {{OpenBMB}},
  year         = {2026},
  howpublished = {\url{https://huggingface.co/openbmb/MiniCPM-RobotManip}}
}

@misc{domino2026,
      title={Towards Generalizable Robotic Manipulation in Dynamic Environments}, 
      author={Heng Fang and Shangru Li and Shuhan Wang and Xuanyang Xi and Dingkang Liang and Xiang Bai},
      year={2026},
      eprint={2603.15620},
      archivePrefix={arXiv},
      primaryClass={cs.CV},
      url={https://arxiv.org/abs/2603.15620}, 
}

@misc{evo1_2025,
      title={Evo-1: Lightweight Vision-Language-Action Model with Preserved Semantic Alignment}, 
      author={Tao Lin and Yilei Zhong and Yuxin Du and Jingjing Zhang and Jiting Liu and Yinxinyu Chen and Encheng Gu and Ziyan Liu and Hongyi Cai and Yanwen Zou and Lixing Zou and Zhaoye Zhou and Gen Li and Bo Zhao},
      year={2025},
      eprint={2511.04555},
      archivePrefix={arXiv},
      primaryClass={cs.RO},
      url={https://arxiv.org/abs/2511.04555}, 
}

@misc{turbovla2026,
      title={TurboVLA: Real-Time Vision-Language-Action Model at 32 Hz on an RTX 4090 with <1 GB VRAM}, 
      author={Hengyi Xie and Chenfei Yao and Xianjin Wu and Yingying Zhu and Dingkang Liang and Xiang Bai and Han Ding},
      year={2026},
      eprint={2607.27205},
      archivePrefix={arXiv},
      primaryClass={cs.CV},
      url={https://arxiv.org/abs/2607.27205}, 
}

@misc{vlaadapter2025,
      title={VLA-Adapter: An Effective Paradigm for Tiny-Scale Vision-Language-Action Model}, 
      author={Yihao Wang and Pengxiang Ding and Lingxiao Li and Can Cui and Zirui Ge and Xinyang Tong and Wenxuan Song and Han Zhao and Wei Zhao and Pengxu Hou and Siteng Huang and Yifan Tang and Wenhui Wang and Ru Zhang and Jianyi Liu and Donglin Wang},
      year={2025},
      eprint={2509.09372},
      archivePrefix={arXiv},
      primaryClass={cs.RO},
      url={https://arxiv.org/abs/2509.09372}, 
}
\bibliographystyle{iclr2027_conference}

\newpage
\appendix

\section{Training Recipes}
\label{app:setup-train}

For every host and every baseline we keep the defaults published with that policy and change only what the comparison requires; the values below are the ones we set. Within a host the \ourmodel{} and vanilla variants share data, initialisation, batch size, learning rate and schedule shape, and differ only by the mechanism and by the number of steps trained, which is set as described under \emph{Training budget} below. Every reported checkpoint is the best one on a held-out validation split, the last $1\%$ of the training data, rather than the last step, under the same selection rule for all variants.

\paragraph{Settings shared by all host variants.} On \textsc{LIBERO} and \textsc{LIBERO-Plus} the chunk is $H{=}50$ with $K{\times}G = 5{\times}10$ and boundary grid $[0,10,20,30,40,50]$; on \textsc{DOMINO} it is $H{=}75$ with $5{\times}15$ and grid $[0,15,30,45,60,75]$. The non-executed loss weight is $\lambda_{\mathrm{ne}}{=}0.25$, and the future-prediction and latent-distillation weights are both $0.1$. All hosts keep an exponential moving average of the weights with decay $0.999$ and warmup $1500$, and every reported score is measured with the EMA weights.

\paragraph{LaWAM.} QwenVL backbone with a $16$-layer flow-matching DiT expert. Both variants initialise from the same pretrained non-streaming checkpoint, so the budget sweep of Sec.~\ref{sec:exp:budget} measures adaptation rather than training from scratch. Global batch $64$; AdamW with betas $(0.9, 0.95)$, $\epsilon{=}10^{-8}$, weight decay $10^{-8}$ and gradient clipping $1.0$; peak learning rate $10^{-4}$, shared by the backbone, the action head and the world model, cosine-decayed to $5\times10^{-7}$ after $1500$ warmup steps; bf16; images at $256$\,px; seed $2026$. The vanilla variant disables the closed-loop path entirely and sees only the chunk endpoints. The decay horizon, the number of steps actually trained and the reported checkpoint differ per benchmark (Table~\ref{tab:app-lawam-steps}).

\begin{table}[htbp]
\centering
\small
\caption{Training-step budgets for the LaWAM host.}
\label{tab:app-lawam-steps}
\begin{tabular}{lrrr}
\toprule
& LIBERO & LIBERO-Plus & DOMINO \\
\midrule
Cosine horizon & 25k & 25k & 50k \\
\midrule
\ourmodel{}, steps trained & 10k & 10k & 20k \\
\ourmodel{}, reported checkpoint & 5k & 8k & 20k \\
\midrule
Vanilla, steps trained & 25k & 25k & 50k \\
Vanilla, reported checkpoint & 20k & 25k & 50k \\
\bottomrule
\end{tabular}
\end{table}

\paragraph{Training budget.} \ourmodel{} is trained for fewer steps than vanilla because its step is more expensive. On \textsc{LIBERO} and \textsc{LIBERO-Plus} the closed-loop step costs $5.862$ TFLOPs per sample against the vanilla step's $3.137$, a ratio of $1.869$; on \textsc{DOMINO}, with three camera views and a $75$-action chunk, the ratio is $1.682$ ($6.826$ against $4.059$). These are measured with a dispatch-layer counter rather than estimated from parameter counts; the counter records zero for elementwise operations, normalisation and softmax, so the absolute values are low by an unknown margin while the ratio is safe. The step counts in Table~\ref{tab:app-lawam-steps} are chosen so that \ourmodel{} remains the cheaper arm on total training FLOPs and on wall-clock, not only on steps (Table~\ref{tab:app-train-budget}). No training-FLOPs measurement exists for the other two hosts, so this parity argument is quantitative on LaWAM and qualitative elsewhere.

\begin{table}[htbp]
\centering
\small
\caption{Total training cost on the LaWAM backbone.}
\label{tab:app-train-budget}
\begin{tabular}{lrrrrrr}
\toprule
& \multicolumn{3}{c}{Training FLOPs (EFLOPs)} & \multicolumn{3}{c}{GPU-hours} \\
\cmidrule(lr){2-4}\cmidrule(lr){5-7}
& \ourmodel{} & Vanilla & Ratio & \ourmodel{} & Vanilla & Ratio \\
\midrule
LIBERO & 3.75 & 5.02 & $0.75\times$ & 11.4 & 12.9 & $0.89\times$ \\
LIBERO-Plus & 3.75 & 5.02 & $0.75\times$ & 11.6 & 12.8 & $0.90\times$ \\
DOMINO & 8.74 & 12.99 & $0.67\times$ & 37.0 & 47.4 & $0.78\times$ \\
\bottomrule
\end{tabular}
\end{table}

\paragraph{$\pi_{0.5}$ and FLOWER.} Everything not listed in Table~\ref{tab:app-other-hosts} follows each policy's own codebase unchanged, and the method-side hyper-parameters are identical to LaWAM's. On FLOWER the \ourmodel{} variant adds the world-model block, which that host does not otherwise have.

\begin{table}[htbp]
\centering
\small
\caption{Training settings for the two transfer hosts. Unlisted settings follow the published defaults.}
\label{tab:app-other-hosts}
\begin{tabular}{@{}lp{0.30\textwidth}p{0.36\textwidth}@{}}
\toprule
& $\pi_{0.5}$ & FLOWER \\
\midrule
Initialisation & released \texttt{pi05\_base} & Florence-2-large with the official $360$k-step pretrained weights \\
Global batch & $8$ & $32$ \\
Steps & $30$k & $30$k ($30$ epochs of $1000$) \\
Reported checkpoint & ours $10$k, vanilla $30$k & ours epoch $10$ ($\approx\!10$k), vanilla epoch $30$ \\
Learning rate & $5\times10^{-5}\!\rightarrow\!5\times10^{-6}$ cosine, $1500$ warmup & $2\times10^{-5}$, the host's three-stage schedule, weight decay $0.05$ \\
Optimiser & AdamW, gradient clipping $1.0$ & AdamW, betas $(0.9, 0.99)$ \\
Precision, images & bf16, $256$\,px & mixed bf16, trained at $112$\,px (evaluation renders at $256$\,px) \\
Chunk $H$ & $50$ & $50$ \\
\bottomrule
\end{tabular}
\end{table}

\paragraph{Compact baselines.} TurboVLA, VLA-Adapter and Evo-1 are trained by us under the \textsc{LIBERO} settings published with each model, unchanged, and with the same settings on all three benchmarks. Every compact-model number in this paper is therefore a reproduction rather than a value copied from the original paper.

\paragraph{Ablations and real-robot runs.} The $K$ and $G$ sweep of Sec.~\ref{sec:disc:kgran} and the real-robot experiments reuse the LaWAM \textsc{LIBERO} \ourmodel{} recipe above and change only the global batch, to $32$ and $16$ respectively. The sweep varies $G \in \{3,5,10,15,20\}$ at $K{=}5$ and $K \in \{3,5,7,10\}$ at $G{=}10$, with the chunk length following as $H = K \times G$; the two sweeps meet at $K{=}5$, $G{=}10$, which is the main recipe. The execution-length curves of Fig.~\ref{fig:exec} and Fig.~\ref{fig:exec-suite} use global batch $64$ on LaWAM and $16$ on $\pi_{0.5}$.

\section{Speed-Measurement Protocol}
\label{app:setup-speed}

We report inference latency, executed-action throughput, and time to first action (TTFA), which capture complementary aspects of inference efficiency. Throughput is computed from the median ($p50$) blocking inference latency, $H_{\mathrm{exec}}/L_{p50}$. TTFA measures the interval from receiving an observation to producing the first executable action; for non-streaming methods it equals the blocking inference latency, while for \ourmodel{} it can be shorter, because the first sub-chunk is emitted before the remaining actions finish denoising.

All measurements use batch $1$ on a single exclusively held device in eager mode with no compilation, CUDA graphs or quantisation. Timing starts from preprocessing and ends when the corresponding actions are available to the controller, excluding model loading and simulator stepping. We synchronise before and after timing, discard $20$ warmup iterations and use at least $200$ timed iterations. Run-to-run spread is about $\pm5\%$, so entries within that band are not ordered.

\subsection{Optimization-Ladder Measurements}
\label{app:speed-ladder}

The rows of Table~\ref{tab:speed} are single blocking chunk calls measured sequentially on one exclusively held GPU. Each cell is $n{=}5$ repeats and we report the median; $\pm$ is one standard deviation.

Both optimizations are bit-exact with the eager implementation, so they change how fast the actions are produced but not the actions themselves.

\subsection{Backbone-Transfer Measurements}
\label{app:speed-hosts}

The speed columns of Table~\ref{tab:hosts} follow the protocol above with two additional constraints, so that the comparison is not confounded by measurement conditions.

\textbf{Both variants are measured together.} For a given backbone, \ourmodel{} and both vanilla configurations are timed in the same session on the same exclusively held device, interleaved rather than run on different days, and the machine load is recorded at the start and end of every run. Neither variant is compiled and neither uses CUDA graphs or quantisation, so no acceleration can be present on one side and absent on the other.

\textbf{Each backbone runs at its released default.} We therefore use the within-backbone speedup as the primary measure of transfer across backbones.

Each cell is $n{=}5$ repeats, and the throughput gains over the $H_{\mathrm{exec}}{=}10$ baseline are $3.62\times$ on LaWAM, $3.21\times$ on $\pi_{0.5}$ and $2.74\times$ on FLOWER.

\section{Full LIBERO Speed and Success Rate}
\label{app:libero}

This section gives the protocol and the numbers behind Fig.~\ref{fig:teaser}. Success rates are as reported by each source paper; we did not reproduce them. Every speed is our own measurement of the released checkpoint at batch $1$ on the same device following Appendix~\ref{app:setup-speed}. Each baseline runs under its own published default inference pipeline and \ourmodel{} in eager mode. Three default pipelines, marked $^{\dagger}$ in Table~\ref{tab:libero}, enable \texttt{torch.compile} or CUDA-graph capture; in eager mode those checkpoints reach $24.9$, $22.2$ and $27.9$ actions per second.

Figure~\ref{fig:teaser} starts its $y$ axis at $94.9\%$; the five policies below that threshold appear in Table~\ref{tab:libero} but not in the figure.

\begin{table}[htbp]
\centering
\caption{LIBERO success rate and measured throughput across methods.}
\label{tab:libero}
\setlength{\tabcolsep}{4pt}
\fitwidth{%
\begin{tabular}{lrrrrrrrr}
\toprule
Method & Spatial & Object & Goal & Long & Avg. & $H_{\mathrm{exec}}$ & Act./s & VRAM (GB) \\
\midrule
\textbf{\ourmodel{} (LaWAM)} & 98.6 & 100.0 & 97.0 & 95.0 & 97.7 & 50 & 292.7 & 5.2 \\
MiniCPM-RobotManip~\citep{minicpm2026robotmanip} & -- & -- & -- & -- & 97.5 & 30 & 166.8 & 3.7 \\
JEPA-WAM~\citep{lin2026jepawam} & 95.6 & 99.4 & 97.2 & 94.6 & 96.7 & 20 & 150.3 & 4.3 \\
Evo-Depth~\citep{lin2026evodepth} & 95.6 & 99.2 & 95.6 & 91.3 & 95.4 & 50 & 145.5 & 3.0 \\
$\pi_{0.5}$~\citep{intelligence2025pi05}$^{\dagger}$ & 98.8 & 98.2 & 98.0 & 92.4 & 96.9 & 5 & 132.1 & 9.5 \\
OpenVLA-OFT~\citep{kim2025openvlaoft} & 97.6 & 98.4 & 97.9 & 94.5 & 97.1 & 8 & 114.7 & 16.1 \\
FLOWER~\citep{reuss2025flower} & 97.5 & 99.1 & 96.1 & 94.9 & 96.9 & 10 & 91.3 & 4.0 \\
VLA-JEPA~\citep{sun2026vlajepa} & 96.2 & 99.6 & 97.2 & 95.8 & 97.2 & 7 & 85.1 & 6.3 \\
InternVLA-M1~\citep{chen2025internvlam1} & 98.0 & 99.0 & 93.8 & 92.6 & 95.9 & 8 & 60.1 & 8.7 \\
BitVLA~\citep{wang2025bitvla} & 96.6 & 99.0 & 95.4 & 92.8 & 96.0 & 8 & 51.2 & 6.6 \\
Fast-WAM~\citep{yuan2026fastwam}$^{\dagger}$ & 98.2 & 100.0 & 97.0 & 95.2 & 97.6 & 10 & 50.1 & 50.0 \\
VLANeXt~\citep{wu2026vlanext} & 99.0 & 99.2 & 96.6 & 94.8 & 97.4 & 8 & 44.0 & 7.8 \\
A1~\citep{zhang2026a1} & 97.4 & 100.0 & 97.4 & 91.0 & 96.5 & 8 & 17.1 & 36.2 \\
Motus~\citep{bi2025motus} & 96.8 & 99.8 & 96.6 & 97.6 & 97.7 & 16 & 8.2 & 33.3 \\
UniVLA~\citep{bu2025univla} & 96.5 & 96.8 & 95.6 & 92.0 & 95.2 & 1 & 7.8 & 15.7 \\
\midrule
$\pi_0$~\citep{black2024pi0}$^{\dagger}$ & 96.8 & 98.8 & 95.8 & 85.2 & 94.2 & 5 & 147.1 & 9.1 \\
SmolVLA~\citep{shukor2025smolvla} & 90.0 & 96.0 & 92.0 & 71.0 & 87.3 & 50 & 280.3 & 1.0 \\
NORA~\citep{hung2025nora} & 92.2 & 95.4 & 89.4 & 74.6 & 87.9 & 5 & 21.2 & 7.7 \\
SpatialVLA~\citep{qu2025spatialvla} & 88.2 & 89.9 & 78.6 & 55.5 & 78.1 & 1 & 2.1 & 8.4 \\
OpenVLA~\citep{kim2024openvla} & 84.7 & 88.4 & 79.2 & 53.7 & 76.5 & 1 & 6.4 & 15.5 \\
\bottomrule
\end{tabular}}
\end{table}

\section{Per-Suite Execution-Length Curves}
\label{app:exec}

\begin{figure}[htbp]
\centering
\includegraphics[width=\linewidth]{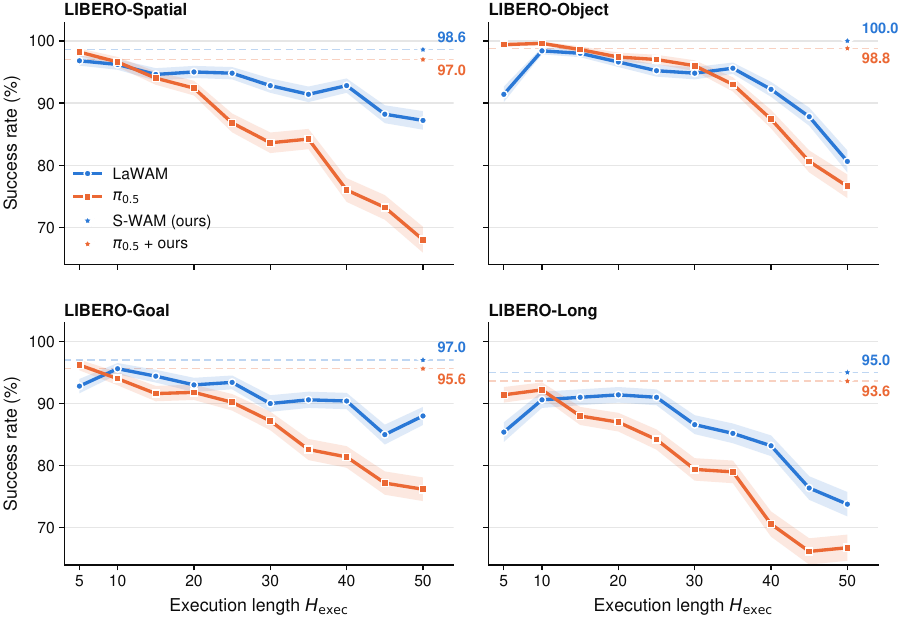}
\caption{Per-suite breakdown of Fig.~\ref{fig:exec}.}
\label{fig:exec-suite}
\end{figure}

Averaged over the four suites our backbone peaks at $H_{\mathrm{exec}}{=}10$ ($95.2$) rather than at the shortest horizon, so success does not fall monotonically as the commitment grows; the per-suite panels of Fig.~\ref{fig:exec-suite} show the effect is carried by LIBERO-Object and LIBERO-Long. The same non-monotonicity appears on hardware in Sec.~\ref{sec:exp:real}, where the vanilla policy is worse at $H_{\mathrm{exec}}{=}20$ than at $50$ on every dynamic task. The checkpoints and global batches behind both figures are given in Appendix~\ref{app:setup-train}.

\section{Detailed Results for the Small-Model Comparison}
\label{app:detailed-results}

Figure~\ref{fig:small-model-compare} summarizes the comparison between \ours{}, two execution-length
variants of the vanilla policy, and three compact-model baselines. This section reports the detailed
numbers behind that figure; the compact models' training settings are given in Appendix~\ref{app:setup-train}.

\subsection{LIBERO}
\label{app:libero-detailed}

Table~\ref{tab:app-libero} reports the per-suite breakdown on LIBERO.

\begin{table}[htbp]
\centering
\small
\caption{Detailed results on LIBERO. Avg.\ is the unweighted mean over the four suites.}
\label{tab:app-libero}
\begin{tabular}{lccccc}
\toprule
Method & Spatial & Object & Goal & Long & Avg. \\
\midrule
TurboVLA & 97.4 & 99.4 & 96.2 & 93.2 & 96.5 \\
VLA-Adapter & 96.6 & 99.8 & 95.8 & 84.0 & 94.0 \\
Evo-1 & 92.8 & 98.2 & 92.4 & 86.4 & 92.5 \\
\midrule
Vanilla, $H_{\mathrm{exec}}=10$ & 96.2 & 98.4 & 95.6 & 90.6 & 95.2 \\
Vanilla, full chunk ($H=50$) & 87.2 & 80.6 & 88.0 & 73.8 & 82.4 \\
\textbf{Ours, full chunk ($H=50$)} & \textbf{98.6} & \textbf{100.0} & \textbf{97.0} & \textbf{95.0} & \textbf{97.7} \\
\bottomrule
\end{tabular}
\end{table}

\subsection{LIBERO-Plus}
\label{app:libero-plus-detailed}

Table~\ref{tab:app-libero-plus} reports the per-category results on LIBERO-Plus. Each entry is first
averaged over the four LIBERO suites, so that a suite does not carry more weight simply because the
benchmark contains more of its episodes. The resulting category-wise unweighted mean (Avg.) is the
number used in the main paper.

\begin{table}[htbp]
\centering
\caption{Per-category LIBERO-Plus results averaged over the four LIBERO suites. Avg.\ is the unweighted mean of the seven categories.}
\label{tab:app-libero-plus}
\fitwidth{%
\begin{tabular}{lrrrrrrrr}
\toprule
Method & Background & Robot & Camera & Language & Noise & Layout & Light & Avg. \\
\midrule
TurboVLA & 77.1 & 30.3 & 76.0 & 72.2 & 71.2 & 60.0 & 81.5 & 66.9 \\
VLA-Adapter & 89.0 & 38.4 & 89.3 & 66.6 & 91.9 & 73.1 & 87.5 & 76.5 \\
Evo-1 & 92.6 & 37.7 & 87.2 & 61.1 & 89.7 & 60.9 & 91.7 & 74.4 \\
\midrule
Vanilla, $H_{\mathrm{exec}}{=}50$ & 83.0 & 49.7 & 77.1 & 53.9 & 80.0 & 64.8 & 85.7 & 70.6 \\
Vanilla, $H_{\mathrm{exec}}{=}10$ & 95.2 & 73.4 & \textbf{91.5} & 77.7 & \textbf{94.3} & \textbf{82.1} & 96.0 & 87.2 \\
\ourmodel{} (ours), $H_{\mathrm{exec}}{=}50$ & \textbf{96.6} & \textbf{76.0} & 89.5 & \textbf{85.2} & 90.4 & 79.3 & \textbf{98.6} & \textbf{87.9} \\
\bottomrule
\end{tabular}}
\end{table}

The seven categories are spread unevenly over the four underlying LIBERO suites, and the ranking
between methods is not the same in every suite. Tables~\ref{tab:app-plus-spatial}--\ref{tab:app-plus-long}
repeat the breakdown one suite at a time. Every cell is recomputed from the per-episode records of the
same runs that produce Table~\ref{tab:app-libero-plus}.
Averaging the Avg.\ columns of the four tables therefore reproduces the Avg.\ column of
Table~\ref{tab:app-libero-plus} and the Avg.\ column of Table~\ref{tab:plus} exactly.

\begin{table}[htbp]
\centering
\caption{Per-category LIBERO-Plus results on \textsc{LIBERO-Spatial}. Avg.\ is the unweighted mean of the seven categories.}
\label{tab:app-plus-spatial}
\fitwidth{%
\begin{tabular}{lrrrrrrrr}
\toprule
Method & Background & Robot & Camera & Language & Noise & Layout & Light & Avg. \\
\midrule
TurboVLA & 91.5 & 28.3 & 69.9 & 82.8 & 68.9 & 54.3 & 92.5 & 69.7 \\
VLA-Adapter & \textbf{98.8} & 50.0 & \textbf{95.7} & 75.4 & \textbf{98.9} & \textbf{93.0} & 98.3 & 87.2 \\
Evo-1 & 92.2 & 36.9 & 87.8 & 68.5 & 91.5 & 63.6 & 89.7 & 75.7 \\
\midrule
Vanilla, $H_{\mathrm{exec}}{=}50$ & 88.0 & 49.4 & 83.0 & 53.3 & 80.1 & 78.2 & 94.2 & 75.2 \\
Vanilla, $H_{\mathrm{exec}}{=}10$ & 98.4 & 73.1 & \textbf{95.7} & 81.5 & 97.2 & \textbf{93.0} & 98.3 & \textbf{91.0} \\
\ourmodel{} (ours), $H_{\mathrm{exec}}{=}50$ & 98.1 & \textbf{75.7} & 93.1 & \textbf{87.2} & 94.3 & 85.2 & \textbf{98.6} & 90.3 \\
\bottomrule
\end{tabular}}
\end{table}

\begin{table}[htbp]
\centering
\caption{Per-category LIBERO-Plus results on \textsc{LIBERO-Object}. Avg.\ is the unweighted mean of the seven categories.}
\label{tab:app-plus-object}
\fitwidth{%
\begin{tabular}{lrrrrrrrr}
\toprule
Method & Background & Robot & Camera & Language & Noise & Layout & Light & Avg. \\
\midrule
TurboVLA & \textbf{99.6} & 36.7 & \textbf{100.0} & \textbf{99.4} & \textbf{99.8} & 78.4 & 99.7 & 87.7 \\
VLA-Adapter & 96.0 & 27.1 & 97.2 & 84.5 & 96.9 & 75.7 & 95.3 & 81.8 \\
Evo-1 & 96.8 & 27.1 & 95.2 & 77.7 & 93.4 & 72.2 & 98.7 & 80.1 \\
\midrule
Vanilla, $H_{\mathrm{exec}}{=}50$ & 84.3 & 35.7 & 78.3 & 58.5 & 84.1 & 65.0 & 89.2 & 70.7 \\
Vanilla, $H_{\mathrm{exec}}{=}10$ & \textbf{99.6} & \textbf{72.1} & 98.5 & 81.1 & 98.6 & \textbf{91.1} & 99.7 & \textbf{91.5} \\
\ourmodel{} (ours), $H_{\mathrm{exec}}{=}50$ & 96.8 & 71.9 & 94.7 & 88.1 & 97.4 & 83.4 & \textbf{100.0} & 90.3 \\
\bottomrule
\end{tabular}}
\end{table}

\begin{table}[htbp]
\centering
\caption{Per-category LIBERO-Plus results on \textsc{LIBERO-Goal}. Avg.\ is the unweighted mean of the seven categories.}
\label{tab:app-plus-goal}
\fitwidth{%
\begin{tabular}{lrrrrrrrr}
\toprule
Method & Background & Robot & Camera & Language & Noise & Layout & Light & Avg. \\
\midrule
TurboVLA & 45.9 & 10.5 & 54.2 & 28.3 & 40.6 & 35.5 & 47.7 & 37.5 \\
VLA-Adapter & 92.5 & 42.5 & \textbf{91.2} & 53.7 & \textbf{93.4} & 59.1 & 82.1 & 73.5 \\
Evo-1 & 91.5 & 39.6 & 85.7 & 44.9 & 85.8 & 50.4 & 92.1 & 70.0 \\
\midrule
Vanilla, $H_{\mathrm{exec}}{=}50$ & 87.5 & 61.9 & 81.9 & 48.3 & 83.9 & 58.8 & 82.1 & 72.1 \\
Vanilla, $H_{\mathrm{exec}}{=}10$ & 94.0 & 78.7 & 90.0 & 67.8 & 92.9 & 63.1 & 90.7 & 82.4 \\
\ourmodel{} (ours), $H_{\mathrm{exec}}{=}50$ & \textbf{95.4} & \textbf{80.0} & 89.5 & \textbf{78.5} & 90.5 & \textbf{67.1} & \textbf{97.8} & \textbf{85.5} \\
\bottomrule
\end{tabular}}
\end{table}

\begin{table}[htbp]
\centering
\caption{Per-category LIBERO-Plus results on \textsc{LIBERO-Long}. Avg.\ is the unweighted mean of the seven categories.}
\label{tab:app-plus-long}
\fitwidth{%
\begin{tabular}{lrrrrrrrr}
\toprule
Method & Background & Robot & Camera & Language & Noise & Layout & Light & Avg. \\
\midrule
TurboVLA & 71.3 & 45.5 & 79.7 & 78.3 & 75.5 & 71.8 & 86.1 & 72.6 \\
VLA-Adapter & 68.5 & 33.8 & 73.0 & 53.0 & 78.4 & 64.7 & 74.5 & 63.7 \\
Evo-1 & 90.0 & 47.3 & 80.1 & 53.3 & 88.4 & 57.4 & 86.1 & 71.8 \\
\midrule
Vanilla, $H_{\mathrm{exec}}{=}50$ & 72.3 & 51.7 & 65.4 & 55.4 & 71.7 & 57.1 & 77.4 & 64.4 \\
Vanilla, $H_{\mathrm{exec}}{=}10$ & 88.9 & 69.5 & \textbf{81.6} & 80.4 & \textbf{88.6} & 81.4 & 95.3 & 83.7 \\
\ourmodel{} (ours), $H_{\mathrm{exec}}{=}50$ & \textbf{96.2} & \textbf{76.3} & 80.7 & \textbf{86.9} & 79.5 & \textbf{81.7} & \textbf{97.8} & \textbf{85.6} \\
\bottomrule
\end{tabular}}
\end{table}

\subsection{DOMINO}
\label{app:domino-detailed}

Table~\ref{tab:app-domino} reports the task-level success rates on DOMINO. Note that DOMINO uses a
75-action chunk, so the ``full chunk'' rows correspond to $H_{\mathrm{exec}}=75$ rather than 50. The
compute-matched baseline for \ours{} on this benchmark is therefore the vanilla full-chunk row.

\begin{table}[htbp]
\centering
\small
\setlength{\tabcolsep}{3pt}
\caption{Task-level success rates (\%) on DOMINO. Avg.\ is the macro-average over the nine tasks.}
\label{tab:app-domino}
\fitwidth{%
\begin{tabular}{lcccccccccc}
\toprule
Method & \shortstack{adjust\\bottle} & \shortstack{beat\\block} & \shortstack{click\\alarm} & \shortstack{click\\bell} & \shortstack{grab\\roller} & \shortstack{move\\can} & \shortstack{move\\card} & \shortstack{press\\stapler} & \shortstack{rotate\\QR} & Avg. \\
\midrule
TurboVLA & 0 & 0 & 8 & 0 & 0 & 0 & 0 & 2 & 0 & 1.11 \\
VLA-Adapter & 10 & 0 & 2 & 0 & 24 & 6 & 0 & 6 & 0 & 5.33 \\
Evo-1 & 0 & 0 & 6 & 0 & 0 & 0 & 0 & 2 & 0 & 0.89 \\
Vanilla, $H_{\mathrm{exec}}=10$ & 0 & 0 & 0 & 0 & 12 & 0 & 0 & 8 & 0 & 2.22 \\
Vanilla, full chunk ($H=75$) & 66 & 6 & 4 & 0 & 30 & 14 & 6 & 14 & 6 & 16.22 \\
\textbf{Ours, full chunk ($H=75$)} & 60 & 24 & 12 & 2 & 38 & 2 & 12 & 20 & 4 & \textbf{19.33} \\
\bottomrule
\end{tabular}}
\end{table}

\subsection{Action Throughput}
\label{app:speed-detailed}

Table~\ref{tab:app-speed-small-models} reports the speed measurements behind the fourth panel of
Fig.~\ref{fig:small-model-compare}. Throughput is reported as executed actions per second
(act/s), computed using the p50 latency of a blocking chunk call in eager mode with batch size 1.

\begin{table}[h]
\centering
\small
\caption{Measured inference efficiency across methods.}
\label{tab:app-speed-small-models}
\begin{tabular}{lccc}
\toprule
Method & Act/s (p50) & TTFA (ms) & VRAM (GB) \\
\midrule
TurboVLA & 340.3 & 35.3 & 0.5 \\
VLA-Adapter & 91.3 & 87.7 & 3.6 \\
Evo-1 & 41.9 & 334.1 & 2.4 \\
Vanilla, $H_{\mathrm{exec}}=10$ & 80.9 & 123.6 & 5.2 \\
Vanilla, $H_{\mathrm{exec}}=50$ & 402.5 & 124.2 & 5.2 \\
\textbf{Ours, $H_{\mathrm{exec}}=50$} & \textbf{292.7} & \textbf{73.3} & 5.2 \\
\bottomrule
\end{tabular}
\end{table}

Overall, these results show that compact-model acceleration can remain competitive on the relatively
standard LIBERO benchmark, but the gap widens on the more challenging LIBERO-Plus and DOMINO
benchmarks. In contrast, \ours{} keeps the larger host policy and improves throughput by amortizing
its inference over a longer execution horizon rather than by reducing model capacity.

\section{Real-Robot Datasets and Detailed Results}
\label{app:real-datasets}

\paragraph{Data collection.}
We collected all real-robot demonstrations on the platform described in
Sec.~\ref{sec:exp:real}, using a 6-DoF UFACTORY xArm~850 with an xArm
Gripper~G2. The robot was teleoperated with a Meta Quest~3, and data was
recorded at $50$\,Hz. The dataset contains nine tasks, with $270$
demonstrations and $305{,}181$ frames in total, corresponding to approximately
$102$ minutes of robot motion. Each demonstration runs until task completion.
For each task, one demonstration is held out for validation and the remaining
demonstrations are used for training.

\paragraph{Observations and actions.}
Each demonstration includes images from a fixed third-person camera and a
wrist-mounted Intel RealSense D435. Both image streams are stored at
$384\times384$. The original $960\times540$ frames are center-cropped to
$540\times540$ and then resized, preserving the aspect ratio. The robot state
is a 7-dimensional vector consisting of the absolute end-effector position,
three Euler angles, and the gripper opening. The action is also
7-dimensional, consisting of the per-step change in end-effector pose and a
binary gripper command, where $1$ indicates closing the gripper.

\paragraph{Tasks.}
Table~\ref{tab:real-datasets} summarizes the nine tasks. Stack bowl, hang cup
and place corn in bowl are each collected in both static and dynamic settings using
the same language instruction. In the dynamic setting, the relevant objects
are placed on a rotating turntable. The remaining tasks cover sustained
manipulation in pour water, precise placement in put lid on the cup, and
multi-stage manipulation in place bowl in drawer.

\begin{table}[t]
  \centering
  \small
  \caption{The nine real-robot datasets. Length denotes the median episode
    duration. Span denotes the range over which manipulated objects are
    re-placed between demonstrations, measured as a percentage of image width.
    All data is recorded at $50$\,Hz.}
  \label{tab:real-datasets}
  \fitwidth{%
  \begin{tabular}{lrrrrl}
    \toprule
    Task & Demos & Frames & Length & Span & Prompt \\
    \midrule
    Stack bowl (static)   & 30 & $27{,}944$ & 19.1\,s & 90\% & \texttt{stack the red bowl on the blue bowl} \\
    Stack bowl (dynamic)  & 30 & $23{,}073$ & 14.6\,s & 69\% & \texttt{stack the red bowl on the blue bowl} \\
    Hang cup (static)     & 30 & $25{,}958$ & 17.0\,s & 52\% & \texttt{hang the yellow cup on the cup rack} \\
    Hang cup (dynamic)    & 30 & $32{,}247$ & 21.2\,s & 49\% & \texttt{hang the yellow cup on the cup rack} \\
    Place corn in bowl (static)  & 30 & $30{,}969$ & 18.1\,s & 99\% & \texttt{place the corn in the bowl} \\
    Place corn in bowl (dynamic) & 30 & $24{,}838$ & 16.1\,s & 78\% & \texttt{place the corn in the bowl} \\
    Pour water            & 30 & $43{,}594$ & 30.2\,s & 93\% & \texttt{pour the water from the red cup into the blue cup} \\
    Put lid on the cup    & 30 & $28{,}456$ & 19.1\,s & 97\% & \texttt{put the blue lid on the blue cup} \\
    Place bowl in drawer  & 30 & $68{,}102$ & 44.4\,s & 92\% & \texttt{put the blue bowl in the second drawer} \\
    \midrule
    Total                 & 270 & $305{,}181$ & & & \\
    \bottomrule
  \end{tabular}}
\end{table}

\paragraph{Scene randomization.}
The robot base and task-specific fixtures remain fixed. Manipulated objects
and distractors are re-placed by hand between demonstrations. The placement
range for each task is reported in Table~\ref{tab:real-datasets}. During
evaluation, objects are re-placed following the same procedure, so the test
scenes follow the same placement distribution as the training demonstrations.

\paragraph{Dynamic tasks.}
The dynamic tasks use a turntable that rotates continuously with a period of
$16$ seconds, corresponding to $22.5^\circ$/s. The turntable rotates
throughout the episode, so the target position changes continuously during
execution. For hang cup, the target orientation changes as well.

\paragraph{Success criteria.}
Table~\ref{tab:real-success} lists the success criterion for each task.

\begin{table}[t]
  \centering
  \small
  \caption{Success criteria for the real-robot evaluation. The static and dynamic
    versions of a task share the same criterion.}
  \label{tab:real-success}
  \begin{tabular}{lp{0.62\linewidth}}
    \toprule
    Task & Counted as a success when \\
    \midrule
    Stack bowl         & the red bowl rests inside the blue bowl after release without tipping or falling outside \\
    Hang cup           & the cup remains on the rack arm after release \\
    Place corn in bowl & the yellow corn is placed inside the bowl and remains there; grasping another object is a failure \\
    Pour water         & the water is poured into the blue cup without dropping the red cup or pouring outside the target \\
    Put lid on the cup & the lid rests on the cup rim and covers the opening without falling onto the table \\
    Place bowl in drawer & the drawer is opened and the bowl is placed inside; both stages are required \\
    \bottomrule
  \end{tabular}
\end{table}

\clearpage

\begin{figure}[p]
  \centering
  \includegraphics[width=\linewidth]{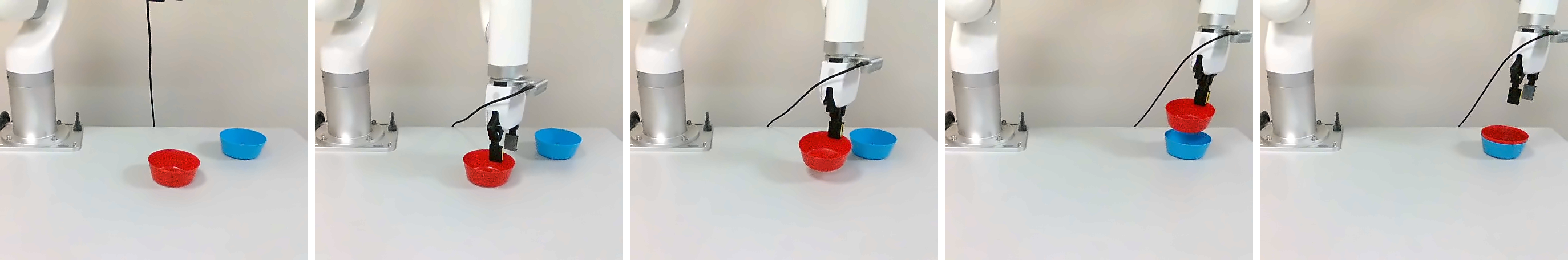}%
  \caption{\textbf{Stack bowl, static.} The robot picks up the red bowl and
    places it inside the blue bowl.}%
  \label{fig:ds-stack-static}
  \vspace{0.6cm}
  \includegraphics[width=\linewidth]{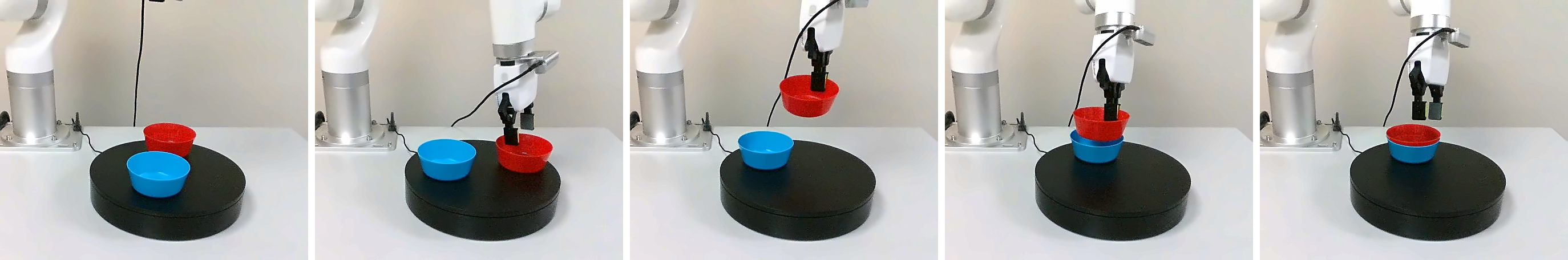}%
  \caption{\textbf{Stack bowl, dynamic.} The same task is performed while both
    bowls rotate on the turntable.}%
  \label{fig:ds-stack-dyn}
  \vspace{0.6cm}
  \includegraphics[width=\linewidth]{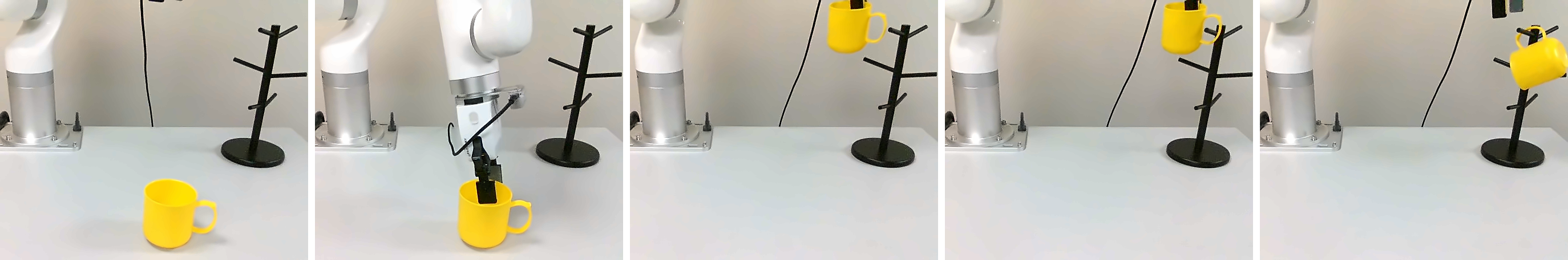}%
  \caption{\textbf{Hang cup, static.} The robot picks up the yellow cup and
    hangs it on the rack.}%
  \label{fig:ds-hang-static}
  \vspace{0.6cm}
  \includegraphics[width=\linewidth]{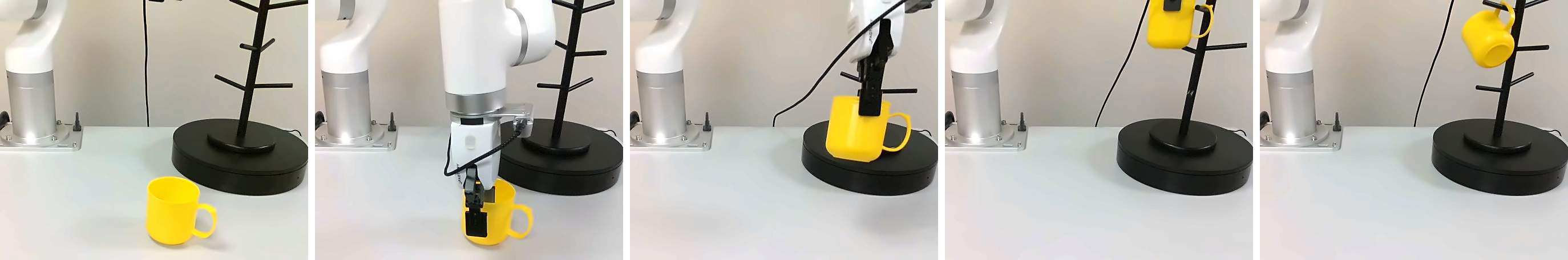}%
  \caption{\textbf{Hang cup, dynamic.} The rack rotates during execution,
    changing both its position and orientation.}%
  \label{fig:ds-hang-dyn}
\end{figure}

\begin{figure}[p]
  \centering
  \includegraphics[width=\linewidth]{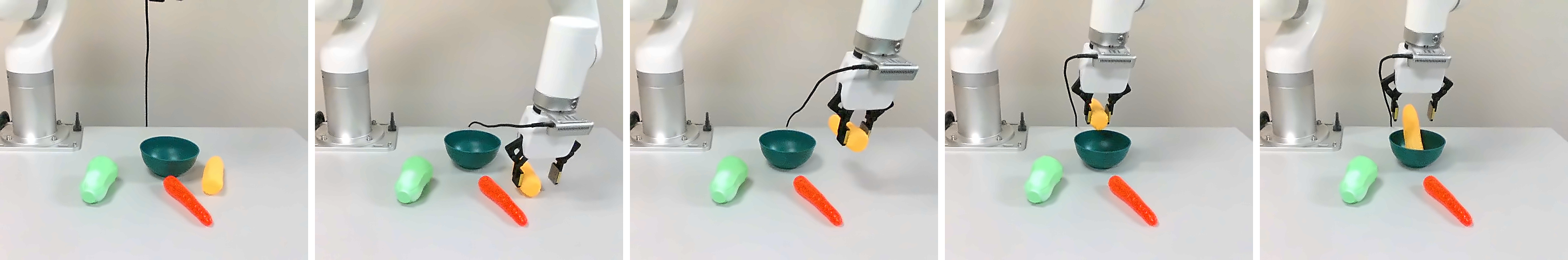}%
  \caption{\textbf{Place corn in bowl, static.} The robot picks up the yellow
    corn and places it in the bowl in the presence of distractor objects.}%
  \label{fig:ds-corn-static}
  \vspace{0.6cm}
  \includegraphics[width=\linewidth]{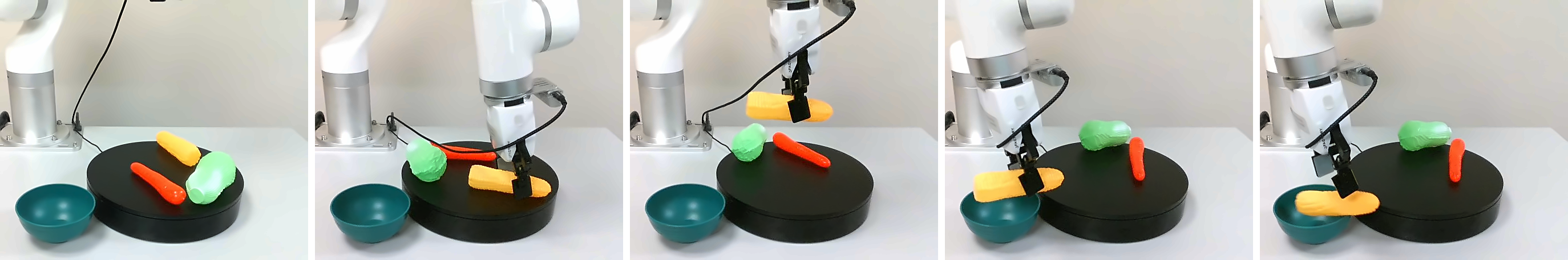}%
  \caption{\textbf{Place corn in bowl, dynamic.} The corn and distractors rotate
    on the turntable while the bowl remains stationary.}%
  \label{fig:ds-corn-dyn}
  \vspace{0.6cm}
  \includegraphics[width=\linewidth]{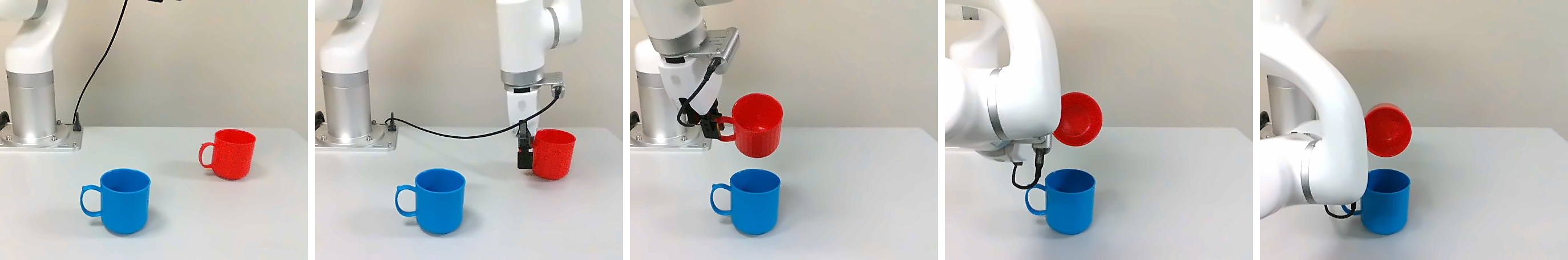}%
  \caption{\textbf{Pour water.} The robot picks up the red cup and pours the
    water into the blue cup.}%
  \label{fig:ds-pour}
  \vspace{0.6cm}
  \includegraphics[width=\linewidth]{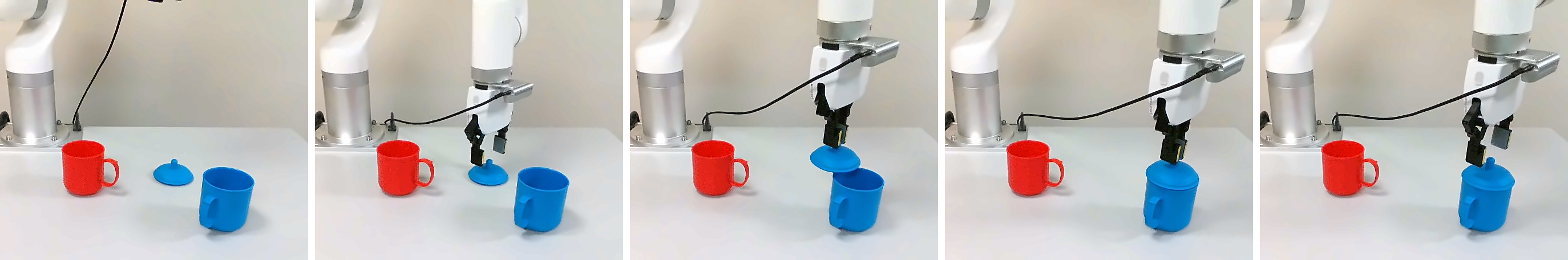}%
  \caption{\textbf{Put lid on the cup.} The robot picks up the blue lid and
    places it on the blue cup.}%
  \label{fig:ds-lid}
  \vspace{0.6cm}
  \includegraphics[width=\linewidth]{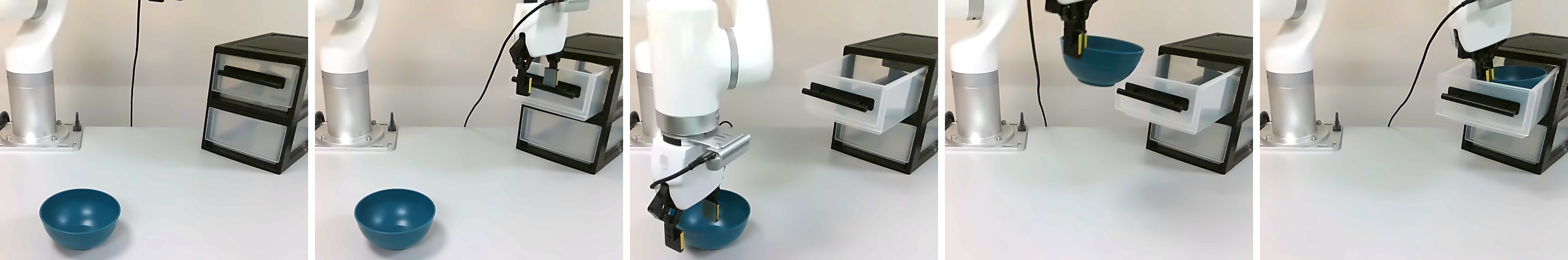}%
  \caption{\textbf{Place bowl in drawer.} The robot first opens the drawer and
    then places the blue bowl inside. The second frame shows the drawer-opening
    stage rather than the object transfer.}%
  \label{fig:ds-drawer}
\end{figure}

\clearpage

\paragraph{Detailed results.}
Table~\ref{tab:app-real} gives the per-task success rates behind Fig.~\ref{fig:real}.
Every task is evaluated over $40$ trials. Static and dynamic averages are the
unweighted means over the six static and the three dynamic tasks respectively.

\begin{table}[t]
  \centering
  \caption{Real-robot success rate (\%) per task, each cell over $40$ trials.
    \ourmodel{} and vanilla at $H_{\mathrm{exec}}{=}50$ are compute-matched;
    vanilla at $H_{\mathrm{exec}}{=}20$ replans $2.5\times$ as often.}
  \label{tab:app-real}
  \begin{tabular}{llrrr}
    \toprule
    Task & Setting & Vanilla, $H_{\mathrm{exec}}{=}20$ & Vanilla, $H_{\mathrm{exec}}{=}50$ & \ourmodel{} \\
    \midrule
    Stack bowl           & Static  & 85.0 & 72.5 & \textbf{90.0} \\
    Hang cup             & Static  & 57.5 & 52.5 & \textbf{70.0} \\
    Place corn in bowl   & Static  & \textbf{77.5} & 60.0 & 75.0 \\
    Put lid on the cup   & Static  & 75.0 & 67.5 & \textbf{80.0} \\
    Pour water           & Static  & 62.5 & 50.0 & \textbf{67.5} \\
    Place bowl in drawer & Static  & 47.5 & 40.0 & \textbf{52.5} \\
    \addlinespace
    \multicolumn{2}{l}{\emph{Static average}} & 67.5 & 57.1 & \textbf{72.5} \\
    \midrule
    Stack bowl           & Dynamic & 10.0 & 25.0 & \textbf{57.5} \\
    Hang cup             & Dynamic & 7.5 & 20.0 & \textbf{27.5} \\
    Place corn in bowl   & Dynamic & 15.0 & 32.5 & \textbf{47.5} \\
    \addlinespace
    \multicolumn{2}{l}{\emph{Dynamic average}} & 10.8 & 25.8 & \textbf{44.2} \\
    \midrule
    \multicolumn{2}{l}{\emph{Overall average}} & 48.6 & 46.7 & \textbf{63.1} \\
    \bottomrule
  \end{tabular}
\end{table}

\end{document}